\documentclass[11pt]{article}
\RequirePackage[colorlinks,citecolor=blue,urlcolor=blue,linkcolor=blue]{hyperref}
\usepackage{url}
\usepackage{booktabs}
\usepackage{svg}
\usepackage{tcolorbox}
\usepackage{subcaption}
\usepackage{url,amscd,paralist,bbm,wrapfig}
\usepackage{bm}
\usepackage{algorithm}
\usepackage{algorithmic} 
\usepackage{todonotes}
\usepackage{amsmath}
\usepackage{amsthm}
\usepackage{amsopn}
\usepackage{amsfonts}
\usepackage{cleveref}

\newcommand*{\email}[1]{\texttt{#1}}

\DeclareSymbolFont{slenderlargesymbols}{OMX}{ccex}{m}{n}

\newtheorem{theorem}{Theorem}
\newtheorem{lemma}{Lemma}
\newtheorem{definition}[theorem]{Definition}
\newtheorem{proposition}{Proposition}
\newtheorem{remark}{Remark}

\usepackage{macro}

\begin{document}
\title{Riemannian AmbientFlow: Towards Simultaneous Manifold Learning and Generative Modeling from Corrupted Data\thanks{Our code is available at \href{https://github.com/wdiepeveen/Riemannian-AmbientFlow}{https://github.com/wdiepeveen/Riemannian-AmbientFlow}.}}
\date{}

\author{Willem Diepeveen\thanks{Department of Mathematics, University of California, Los Angeles (email: \email{wdiepeveen@math.ucla.edu})}
\and Oscar Leong\thanks{Department of Statistics and Data Science, University of California, Los Angeles (email: \email{oleong@stat.ucla.edu})} 
}

\maketitle
\begin{abstract}
    Modern generative modeling methods have demonstrated strong performance in learning complex data distributions from clean samples. In many scientific and imaging applications, however, clean samples are unavailable, and only noisy or linearly corrupted measurements can be observed. Moreover, latent structures, such as manifold geometries, present in the data are important to extract for further downstream scientific analysis. In this work, we introduce Riemannian AmbientFlow, a framework for simultaneously learning a probabilistic generative model and the underlying, nonlinear data manifold directly from corrupted observations. Building on the variational inference framework of AmbientFlow, our approach incorporates data-driven Riemannian geometry induced by normalizing flows, enabling the extraction of manifold structure through pullback metrics and Riemannian Autoencoders. We establish theoretical guarantees showing that, under appropriate geometric regularization and measurement conditions, the learned model recovers the underlying data distribution up to a controllable error and yields a smooth, bi-Lipschitz manifold parametrization. We further show that the resulting smooth decoder can serve as a principled generative prior for inverse problems with recovery guarantees. We empirically validate our approach on low-dimensional synthetic manifolds and on MNIST.
\end{abstract}

\section{Introduction}

Uncovering latent structure from corrupted or incomplete data is a long-standing challenge in modern data science and machine learning. In many applications, data of interest may not be able to be inferred directly, and the acquisition process may be hindered by noise, missing measurements, or systematic corruption. Addressing such challenges have important applications across diverse fields, such as medical imaging \cite{lustig2007sparse},  astronomical imaging \cite{chael2019eht}, geophysics \cite{virieux2009overview}, and more. The underlying data signals are typically highly structured, allowing the use of priors or regularizers to effectively model such data aiding in signal recovery.

A particularly effective structural model is the manifold hypothesis: high-dimensional natural data often concentrate near a low-dimensional, nonlinear manifold \cite{fefferman2016testing}. A recent line of work has enabled one to efficiently extract such data manifolds -- along with intrinsic dimensionality, geodesics, and other manifold mappings -- under learned pullback Riemannian structures \cite{diepeveen2025scorebased,diepeveen2025manifold} induced by normalizing flow-based generative models \cite{dinh2014nice,dinh2017density,kingma2018glow}. Beyond aiding algorithms for clustering, classification, or anomaly detection, explicit access to such geometric quantities and attributes also improves interpretability \cite{diepeveen2025iso} and enables sample-efficient learning \cite{kruiff2025pullback}. Yet, most of these works typically presume access to clean samples.


In parallel, there has been rapid progress in the machine learning community on learning generative models from corrupted data alone. Several paradigms for training such models include those based on Generative Adversarial Networks (GANs) \cite{bora2018ambientgan, kabkab2018task}, normalizing flow models \cite{kelkarambientflow}, Variational Autoencoders (VAEs) \cite{olsen2022data, mendoza2022self}, and diffusion models \cite{daras2023ambientdiffusion, weiminbai2024emdiffusion, daras2024ambienttweedie, chen2025denoising}. This is motivated by applications in which ground-truth data may be expensive, difficult to observe, or impossible to obtain, but it is still desirable to obtain an approximate prior of the underlying data distribution\footnote{Additionally, training generative models on corrupted data has shown to reduce memorization of the training set \cite{shah2025does}.} for downstream tasks. For example, once inferred, such models can be exploited to solve downstream tasks such as inverse problems \cite{leong2023discovering, lin2024imaging, aali2025ambient-mri, zhang2025restoration}. 

Although such methods have shown strong generation performance in the presence of noisy data, they are inherently limited in their ability to describe the underlying structure in the data. In particular, pure generative models typically do not explicitly capture geometric features such as intrinsic dimensionality, geodesics, or other manifold mappings, which are essential for meaningful downstream tasks such as clustering, classification, anomaly detection, and interpretability.


Based on the above threads, we consider the following question:

\vspace{0.2cm}
\begin{center}
    \textit{Can we simultaneously recover the underlying geometric structure (e.g., a data manifold) and learn a probabilistic generative model from corrupted samples of a data distribution?}
\end{center}
\vspace{0.2cm}


In this paper, we present a framework that combines generative modeling from corruption with Riemannian pullback geometry to simultaneously identify manifold structure and learn a probabilistic generative model. Our main contributions are as follows:
\begin{enumerate}
    \item We present \emph{Riemannian AmbientFlow}, a framework that
    integrates variational methods for learning from noisy data \cite{kelkarambientflow} with Riemannian geometric methods for manifold learning \cite{diepeveen2025scorebased,diepeveen2025manifold}. Our method can recover the correct data distribution from indirect, noisy measurements, with a controllable error (\cref{thm:recoverability}), which directly generates a Riemannian Autoencoder (RAE) for geometric analysis and sampling, with explicit reconstruction error bounds (\cref{thm:expected-proj-err}). 
    \item We highlight several benefits of the RAE, such as showing that the RAE decoder is bi-Lipschitz and smooth under suitable parametrization (\cref{prop:decoder-smoothness}). Moreover, we showcase its applicability as a prior for inverse problems. In particular, when the underlying measurement operator satisfies an isometry property over the range of the RAE decoder, gradient descent converges linearly to the true signal in the range of the decoder (\cref{thm:gd-convergence-ips}). 
    \item We empirically validate our approach on several datasets, including low-dimensional synthetic manifolds and MNIST, showing that it is possible to jointly learn the underlying manifold geometry and a generative model directly from corrupted samples.
\end{enumerate}


\subsection{Related Work} 



\paragraph{Riemannian Manifold Learning}

Despite the focus of this work being on using pullback geometry for Riemannian manifold learning, it is worthwhile to discuss how constructing a Riemannian structure to extract the data manifold -- through non-linear low-rank approximation \cite{fletcher2004principal} whilst accounting for curvature \cite{diepeveen2025curvature} -- can be achieved in various ways. However, early efforts that attempted to directly construct metric tensor fields \cite{arvanitidis2016locally,hauberg2012geometric,peltonen2004improved} were not very scalable on real (high-dimensional) data, both in terms of learning the geometry in the first place and evaluating manifold mappings subsequently. Later approaches introduced specialized structures to improve scalability for learning the geometry, either by making assumptions about the data \cite{Scarvelis2023} or about the Riemannian structure itself \cite{sorrenson2025learning}. However, scalability of manifold mappings remained a challenge. The first step here was made through learning a pullback structure \cite{diepeveen2024pulling} derived from pairwise distances \cite{fefferman2020reconstruction}. Such Riemannian manifolds come with closed-form manifold mappings and stability guarantees when the pullback structure is a local $\ell^2$-isometry on the data support. However, in \cite{diepeveen2024pulling} the authors also faced a limitation in their proposed approach, specifically its lack of scalability when learning the geometry. Only recently, this gap between scalable training and scalable manifold mappings was bridged in \cite{diepeveen2025scorebased}, in which the authors demonstrated that a diffeomorphism obtained through adapted normalizing flow training \cite{dinh2017density} is an effective method for constructing pullback geometry. Notably, by restricting the expressivity of the normalizing flows even further \cite{dinh2014nice,kingma2018glow}, regular normalizing flow training actually suffices \cite{diepeveen2025manifold}.

\paragraph{Ambient Distribution Learning} Learning high-quality generative models is typically limited by the availability of large, clean datasets of images of interest. This can be particularly challenging in scientific applications, where clean data may either be time-consuming, expensive, or impossible to obtain, which includes areas such as astronomy~\cite{roddier1988interferometric, lin2024imaging}, medical imaging~\cite{reed2021dynamic, jalal2021robust}, and seismology~\cite{nolet2008breviary, rawlinson2014seismic}. For instance, fully sampled MRI acquisitions are time-consuming and uncomfortable for patients~\cite{knoll2020fastmri}. Moreover, classical hand-crafted priors are typically less flexible and can suffer from human bias \cite{levin2009understanding}. Hence there has been significant interest in developing methods that can either directly solve the underlying inverse problem by learning an implicit prior \cite{akyildiz2025efficient, leong2023discovering}, exploiting equivariance \cite{chen2023imaging}, or by developing generative modeling methods to directly learn generative models from corrupted data. The latter class of methods fall under the \textit{Ambient Distribution Learning} paradigm, where ambient refers to the fact that measurement examples are leveraged \cite{daras2023ambientdiffusion, daras2025ambientscaling, aali2025ambient-mri, weiminbai2024emdiffusion}. One of the earliest works along these lines was AmbientGAN \cite{bora2018ambientgan}, which learned a Generative Adversarial Network \cite{goodfellow2020generative} directly from noisy measurements by enforcing that the induced measurement distribution mimicked the given measurement distribution. Several subsequent methods have been developed for GANs \cite{kabkab2018task}, VAEs \cite{mendoza2022self}, and normalizing flows \cite{kelkarambientflow}. Given the advent of diffusion and score-based models, recent work has shown how additional masking techniques can learn diffusion models from linearly corrupted data \cite{daras2023ambientdiffusion} along with Gaussian noise \cite{daras2024ambienttweedie, daras2025ambientscaling}. Additional works include those exploiting the Expectation-Maximization (EM) algorithm \cite{rozet2024learning, weiminbai2024emdiffusion} and distillation techniques \cite{chen2025denoising, zhang2025restoration}. Recent works have also considered how to address black-box and arbitrary corruptions via diffusion models \cite{daras2025ambient} and those based on stochastic interpolants \cite{modi2025generative}. However, each of these methods focus mainly on the task of learning a probabilistic generative model from the data and do not explicitly exploit any underlying latent (such as Riemannian manifold-based) structure in the data.

\section{Preliminaries}


\subsection{Data-driven Riemannian Geometry}
\label{sec:dd-rg}
For the purposes of this work we will need the following notions and results, which we present in basic notations from differential and Riemannian geometry, see \cite{boothby2003introduction,carmo1992riemannian,lee2013smooth,sakai1996riemannian} for details.

\paragraph{Smooth Manifolds and Tangent Spaces} 
Let $\manifold$ be a \emph{$\dimInd$-dimensional smooth manifold}. We write $C^\infty(\manifold)$ for the space of smooth functions over $\manifold$. The \emph{tangent space} at $\mPoint \in \manifold$, which is defined as the space of all \emph{derivations} at $\mPoint$, is denoted by $\tangent_\mPoint \manifold$ and for \emph{tangent vectors} we write $\tangentVector_\mPoint \in \tangent_\mPoint \manifold$. For the \emph{tangent bundle} we write $\tangent\manifold$ and smooth vector fields, which are defined as \emph{smooth sections} of the tangent bundle, are written as $\vectorfield(\manifold) \subset \tangent\manifold$.

\paragraph{Riemannian Manifolds} 
A smooth manifold $\manifold$ becomes a \emph{Riemannian manifold} if it is equipped with a smoothly varying \emph{metric tensor field} $(\cdot, \cdot) : \vectorfield(\manifold) \times \vectorfield(\manifold) \to C^\infty(\manifold)$. This tensor field induces a \emph{(Riemannian) metric} $\distance_{\manifold} : \manifold\times\manifold\to\R$. The metric tensor can also be used to construct a unique affine connection, the \emph{Levi-Civita connection}, that is denoted by $\nabla_{(\,\cdot\,)}(\,\cdot\,) : \vectorfield(\manifold) \times \vectorfield(\manifold) \to \vectorfield(\manifold)$. 
This connection is in turn the cornerstone of a myriad of manifold mappings.

One is the notion of a \emph{geodesic}, which for two points $\mPoint,\mPointB \in \manifold$ is defined as a curve $\geodesic_{\mPoint,\mPointB} : [0,1] \to \manifold$ with minimal length that connects $\mPoint$ with $\mPointB$ -- that is, if such a curve exists. Another closely related notion to geodesics is the curve $t \mapsto \geodesic_{\mPoint,\tangentVector_\mPoint}(t)$  for a geodesic starting from $\mPoint\in\manifold$ with velocity $\dot{\geodesic}_{\mPoint,\tangentVector_\mPoint} (0) = \tangentVector_\mPoint \in \tangent_\mPoint\manifold$. This can be used to define the \emph{exponential map} $\exp_\mPoint : \mathcal{D}_\mPoint \to \manifold$ at $\mPoint$ as \(\exp_\mPoint(\tangentVector_\mPoint) := \geodesic_{\mPoint,\tangentVector_\mPoint}(1),\) where \(\mathcal{D}_\mPoint \subset \tangent_\mPoint \manifold\) is the set on which \(\geodesic_{\mPoint,\tangentVector_\mPoint}(1)\) is defined. Furthermore, the \emph{logarithmic map} $\log_\mPoint : \exp_\mPoint(\mathcal{D}'_\mPoint ) \to \mathcal{D}'_\mPoint$ at $\mPoint$ is defined as the inverse of $\exp_\mPoint$, so it is well-defined on  $\mathcal{D}'_{\mPoint} \subset \mathcal{D}_{\mPoint}$ where $\exp_\mPoint$ is a diffeomorphism. 


\paragraph{Pullback Manifolds} 
If $(\manifold, (\cdot,\cdot))$ is a $\dimInd$-dimensional Riemannian manifold, $\manifoldB$ is a smooth $\dimInd$-dimensional manifold and $\diffeo:\manifoldB \to \manifold$ is a diffeomorphism, the \emph{pullback metric}
\begin{equation}
    (\tangentVector, \tangentVectorB)_\mPoint^\diffeo := (D_{\mPoint}\diffeo[\tangentVector_{\mPoint}], D_{\mPoint}\diffeo[\tangentVectorB_{\mPoint}])_{\diffeo(\mPoint)},  \mPoint \in \manifoldB, \tangentVector, \tangentVectorB \in \vectorfield(\manifoldB),
    \label{eq:pull-back-metric}
\end{equation}
where $D_{\mPoint}\diffeo: \tangent_\mPoint \manifoldB \to \tangent_{\diffeo(\mPoint)} \manifold$ denotes the differential of $\diffeo$,
defines a Riemannian structure on $\manifoldB$, which we denote by $(\manifoldB, (\cdot,\cdot)^\diffeo)$. 
Pullback mappings are denoted similarly to (\ref{eq:pull-back-metric}) with the diffeomorphism $\diffeo$ as a superscript, i.e., we write $\distance^\diffeo_{\manifoldB}(\mPoint, \mPointB)$, $\geodesic^\diffeo_{\mPoint, \mPointB}$, $\exp^\diffeo_\mPoint (\tangentVector_\mPoint)$, and $\log^\diffeo_{\mPoint} \mPointB$ for $\mPoint,\mPointB \in \manifoldB$ and $\tangentVector_\mPoint \in \tangent_\mPoint \manifoldB$. Pullback metrics literally pull back all geometric information from the Riemannian structure on $\manifold$. 
In particular, closed-form manifold mappings on $(\manifold, (\cdot,\cdot))$ yield under mild assumptions closed-form manifold mappings on $(\manifoldB, (\cdot,\cdot)^\diffeo)$. 

\paragraph{Data-driven Pullback Manifolds}

Notably, for Euclidean pullback manifolds $(\R^\dimInd,(\cdot,\cdot)^\diffeo)$ generated by a diffeomorphism $\diffeo:\R^\dimInd\to \R^\dimInd$ pulling back the standard Euclidean structure $(\R^\dimInd, (\cdot, \cdot)_2)$ -- which is how scalable data-driven Riemannian geometry is constructed for high-dimensional data \cite{diepeveen2025scorebased,diepeveen2025manifold} --, we have \cite[Prop~2.1]{diepeveen2024pulling}
\begin{align}
    \distance_{\R^{\dimInd}}^{\diffeo}(\Vector, \VectorB) &= \|\diffeo(\Vector) - \diffeo(\VectorB)\|_2,
    \label{eq:thm-distance-remetrized}\\
    \geodesic^{\diffeo}_{\Vector, \VectorB}(t) &= \diffeo^{-1}((1 - t)\diffeo(\Vector) + t \diffeo(\VectorB)),
    \label{eq:thm-geodesic-remetrized}\\
    \exp^{\diffeo}_\Vector (\tangentVector_\Vector) &= \diffeo^{-1}(\diffeo(\Vector) + D_{\Vector} \diffeo[\tangentVector_\Vector]),
    \label{eq:thm-exp-remetrized}\\
    \log^{\diffeo}_{\Vector} (\VectorB) &= D_{\diffeo(\Vector)}\diffeo^{-1}[\diffeo(\VectorB) - \diffeo(\Vector)],
    \label{eq:thm-log-remetrized}
\end{align}
where $\Vector, \VectorB\in \R^\dimInd$ and $\tangentVector_\Vector \in \tangent_\Vector \R^\dimInd \cong \R^\dimInd$, along with \cite[Prop~3.7]{diepeveen2024pulling}
\begin{equation}
    \argmin_{\Vector\in \R^\dimInd} \sum_{\sumIndA=1}^N \distance^\diffeo_{\R^\dimInd}(\Vector, \Vector^\sumIndA)^2 = \diffeo^{-1} (\frac{1}{N} \sum_{\sumIndA=1}^N \diffeo(\Vector^\sumIndA)),
    \label{eq:thm-bary-remetrized}
\end{equation}
for the Riemannian barycentre \cite{karcher1977riemannian}, where $\Vector^1, \ldots, \Vector^N\in \R^\dimInd$. 

In the context of a data-driven pullback structure, the manifold mappings above gain a practical interpretation. A well-trained $\diffeo$ essentially flattens out the data space, i.e., it maps a data set -- residing close to a non-linear data manifold -- into the vicinity of a (low-dimensional) linear subspace of $\R^\dimInd$. Manifold mappings are essentially computed using Euclidean rules applied to points and tangent vectors mapped into this linear subspace by $\diffeo$ and then mapped back to the original data domain using $\diffeo^{-1}$. As a result, geodesics between two points will always move through regions with large amounts of data -- or probabilistically speaking through regions with high likelihood. For a more detailed discussion and the manifold mapping for the general pullback setting, we refer the reader to \cite{diepeveen2024pulling}.

\paragraph{Learning Euclidean Pullback Manifolds}
To learn such a diffeomorphism $\diffeo:\R^\dimInd\to\R^\dimInd$ that generates geodesics that interpolate through regions of high likelihood, normalizing flow training has shown to be a scalable approach \cite{diepeveen2025scorebased,diepeveen2025manifold}. This boils down to minimizing the negative log likelihood loss for variations of a distribution $\density_{\spdMatrix,\networkParams}: \R^\dimInd \to \R$ of the form
\begin{equation}
    \density_{\spdMatrix,\networkParams}(\Vector)\propto e^{-\frac{1}{2} \diffeo_\networkParams(\Vector)^\top \spdMatrix^{-1} \diffeo_\networkParams(\Vector) } ,
    \label{eq:stat-model-geom}
\end{equation}
where $\spdMatrix\in \R^{\dimInd\times \dimInd}$ is a positive definite matrix and the mapping $\diffeo_\networkParams:\R^\dimInd\to\R^\dimInd$ is a diffeomorphic neural network with parameters $\networkParams$. 

To gain intuition as to why this approach yields a suitable diffeomorphism and pullback geometry by extension, we first note that the function $t \mapsto -\log(\density_{\spdMatrix,\networkParams}(\geodesic^{\diffeo_{\networkParams}}_{\Vector, \VectorB}(t)))$ is convex for any combination of end points $\Vector,\VectorB\in \R^\dimInd$ and network parametrization $\networkParams$. Then, if $\density_{\text{data}}$ is feasible, i.e., there exists some $\networkParams^*$ such that $\density_{\networkParams^*} = \density_{\text{data}}$, minimizing the negative log likelihood loss will find this $\networkParams^*$. In other words, if we have $\density_{\networkParams^*} = \density_{\text{data}}$ this means that geodesics $\geodesic^{\diffeo_{\networkParams^*}}_{\Vector, \VectorB}$ between data points move through regions with higher likelihood than the end points, which is exactly what we set out to do.

In reality, we cannot expect the equality $\density_{\networkParams^*} = \density_{\text{data}}$ to hold exactly. Instead, the best we can hope for is that $\density_{\networkParams^*} \approx \density_{\text{data}}$ -- especially if the data is inherently multimodal. Nevertheless, this approach still tends to yield suitable\footnote{for downstream geometric purposes} diffeomorphisms in practice, see \cite{diepeveen2025scorebased,diepeveen2025manifold} for details.

\paragraph{Riemannian Autoencoders} One such downstream tasks is compression through the construction of a \emph{Riemannian Autoencoder} (RAE) \cite{diepeveen2024pulling}. An RAE  is a mapping of the form
\begin{equation}
    \Vector \mapsto (\RAEdecoder \circ \RAEencoder) (\Vector),
\end{equation}
where $\RAEencoder:\R^\dimInd \to \R^{\dimIndC}$ and $\RAEdecoder:\R^{\dimIndC} \to \R^\dimInd$ are defined as
\begin{equation}
    \RAEencoder(\Vector) := \mathbf{U}^\top \log^{\diffeo_\networkParams}_{\bar{\Vector}} (\Vector) \quad \text{and} \quad \RAEdecoder_\RAErelerror(\latentVector):= \exp_{\bar{\Vector}}^{\diffeo_\networkParams} \Bigl( \mathbf{U}\latentVector \Bigr).
    \label{eq:rae-encoder-nf-example}
\end{equation}
Here $\bar{\Vector} \in \R^\dimInd$ is a chosen base point, and $\mathbf{U} := [\mathbf{u}^{1}, \ldots, \mathbf{u}^\dimIndC] \in (\tangent_{\bar{\Vector}} \R^\dimInd)^\dimIndC \cong \R^{\dimInd \times \dimIndC}$ a chosen orthonormal basis in the tangent space at the base point. Both the base point and basis can be chosen in several ways, see \cite{diepeveen2024pulling,diepeveen2025manifold,diepeveen2025scorebased} for different approaches. However, for any of these settings, the RAE mapping can be seen as a non-linear projection onto a $\dimIndC$-dimensional manifold approximating the data set.



\subsection{AmbientFlow}

A method particularly close to our work is AmbientFlow \cite{kelkarambientflow}, which develops an approach based on normalizing flows to learn a distribution from corrupted examples. Suppose there is a ground-truth distribution of images $\stoVector \sim \density_{\text{data}}$ but we only have access to linearly corrupted samples $\stoCorVector \sim \densityB_{\text{data}}$ given by 
$$
\stoCorVector \sim \densityB_{\text{data}} \Longleftrightarrow \stoCorVector = \forward\stoVector + \mathbf{n},\ \stoVector \sim \density_{\text{data}}, \mathbf{n} \sim \density_{\text{noise}},\ \forward \in \R^{\dimIndB \times \dimInd}.
$$ 
Since only measurement examples are given $\stoCorVector \sim \densityB_{\text{data}}$, the authors in \cite{kelkarambientflow} propose to learn a generative model by encouraging the induced measurement distribution $\psi_{\networkParams} := \forward_{\sharp}(\density_{\networkParams}) * \density_{\text{noise}}$ to be close to $\densityB_{\text{data}}$, which can be done by minimizing $D_{\mathrm{KL}}(\densityB_{\text{data}}|| \psi_{\networkParams})$. A challenge, however, is that this quantity is inaccessible due to the need to calculate quantities such as $\log \densityB_{\text{data}}$, leading them to consider (equivalently) minimizing the following objective over an additional set of parameters $(\networkParams, \networkParamsB)$: \begin{align*}
    \mathcal{L}(\theta, \networkParamsB) := \mathbb{E}_{\stoCorVector \sim \densityB_{\text{data}}}\left[\log \mathbb{E}_{\stoVector \sim \density_{\eta}(\cdot \mid \stoCorVector)}\left\{\frac{\density_{\networkParams}(\stoVector) \density_{\text{noise}}(\stoCorVector - \forward\stoVector)}{\density_{\networkParamsB}(\stoVector\mid \stoCorVector)}\right\}\right].
\end{align*} Here, an additional posterior network given by a conditional normalizing flow $\Vector\mapsto \density_{\eta}(\Vector \mid \CorVector)$ is used to generate samples from $\density_{\networkParamsB}(\Vector|\CorVector) \approx \density_{\networkParams}(\Vector|\CorVector).$ To encourage closeness of these posteriors, a variational lower bound is optimized: \begin{multline}
     \mathcal{L}_{\text{VLB}}(\networkParams,\networkParamsB) := \\ \mathbb{E}_{\stoCorVector \sim \densityB_{\textbf{data}}, \stoVector^\sumIndA \sim \density_{\eta}(\cdot \mid \stoCorVector)}\Biggl[\log \biggl( \frac{1}{M}\sum_{\sumIndA=1}^M \exp\bigl[\log \density_{\networkParams}(\stoVector^\sumIndA) + \log \density_{\text{noise}}(\stoCorVector - \forward \stoVector^\sumIndA) - \log\density_{\networkParamsB}(\stoVector^\sumIndA\mid\stoCorVector)\bigr]\biggr)\Biggr].
     \label{eq:vlb-ambientflow}
\end{multline} 

In most applications of interest, the linear map $\forward \in \R^{\dimIndB \times \dimInd}$ has a non-trivial kernel, leading to a loss of information $\density_{\text{data}} \mapsto \densityB_{\text{data}}$. Without further structural assumptions on the underlying data density, it is impossible to expect exact recovery of the underlying data distribution. The authors in \cite{kelkarambientflow} tackle this by assuming that the underlying distribution $\density_{\text{data}}$ and the distributions they search over $\density_{\text{data}}$ concentrate on signals that are compressible with respect to a sparsifying basis. By adding an additional constraint to search over such distributions, they show that global minimizers of the variational lower bound $\mathcal{L}_{\text{VLB}}$ subject to such a constraint are close in Wasserstein-distance to the underlying ground-truth image distribution.

\section{Framework}\label{sec:framework}
In this section we develop \emph{Riemannian AmbientFlow}, which  differs from AmbientFlow \cite{kelkarambientflow} in the type of sparsity we assume. In particular, while still being in the setting of trying to learn a generative model from corrupted data, the main difference is that we assume that this data distribution is concentrated near a low-dimensional manifold that is the range of a Riemannian Autoencoder (RAE) generated by $\diffeo$. In the following we will see that from a distribution learning perspective this is related to assuming that the ground truth data are distributed as 
\begin{equation}
    \density(\Vector) := \frac{1}{\sqrt{(2\pi)^\dimInd \det(\spdMatrix)}}e^{-\frac{1}{2} \diffeo(\Vector)^\top \spdMatrix^{-1} \diffeo(\Vector) } |\det(D_{\Vector} \diffeo)|
    \label{eq:paper-density-model}
\end{equation}
where $\spdMatrix\in \R^{\dimInd\times \dimInd}$ is a positive definite matrix and $\diffeo:\R^d\to\R^d$ is a diffeomorphism such that the mapping $\Vector\mapsto |\det(D_{\Vector} \diffeo)|$ is constant\footnote{So similar, but slightly different to the settings of \cite{diepeveen2025scorebased,diepeveen2025manifold}. Note that the motivation for the constant determinant Jacobian is to ensure that the mapping $t \mapsto -\log(\density(\geodesic^{\diffeo}_{\Vector, \VectorB}(t)))$ is convex for any $\Vector,\VectorB\in \R^\dimInd$.}. 

The main difficulty lies in the absence of ground-truth data for learning \cref{eq:paper-density-model} through anisotropic normalizing flow training, as well as the lack of a canonical choice of base point or orthonormal basis to define a suitable RAE induced by $\diffeo$. Our first step toward addressing this issue is to reformulate the low-dimensionality assumption so that it applies directly to $\spdMatrix$ and $\diffeo$ themselves, without requiring an explicit choice of dimension. This reformulation enables a geometric regularization of the AmbientFlow framework, thereby facilitating joint manifold learning and generative modeling. We refer to this integrated method as \emph{Riemannian AmbientFlow}, and we demonstrate that it admits recovery guarantees.

\subsection{Riemannian Autoencoders from a Known Distribution}
Assuming we know the distribution \cref{eq:paper-density-model}, our first step is to pick an $\dimIndC \ll \dimInd$ and construct a RAE with encoder and decoder $\RAEencoder:\R^\dimInd \to \R^{\dimIndC}$ and $\RAEdecoder:\R^{\dimIndC} \to \R^\dimInd$ generated by $\density$ such that the \emph{distributional projection error} $W_1((\RAEdecoder \circ \RAEencoder)_{\sharp} \density, \density)$ is small. Here, $W_1$ denotes the Wasserstein-1 distance between measures, which is given by $$W_1(p,q):= \inf_{\gamma \in \Gamma(p,q)} \mathbb{E}_{(\Vector,\Vector')\sim\gamma}[\|\Vector - \Vector'\|_2],$$ where $\Gamma(p,q)$ is the set of all couplings between $p$ and $q$. The motivation for this error is that it quantifies the extent to which our assumption of the data distribution being concentrated near an $\dimIndC$-dimensional manifold is satisfied. The following result tells us that it suffices to construct an RAE with small \emph{expected projection error} $\mathbb{E}_{\stoVector\sim \density}\Bigl[ \|\RAEdecoder(\RAEencoder(\stoVector)) - \stoVector\|_2\Bigr]$. 

\begin{proposition}[Distributional vs expected projection error]
\label{prop:mfld-dist-error-proj}
Let $\RAEencoder:\R^\dimInd \to \R^{\dimIndC}$ and $\RAEdecoder:\R^{\dimIndC} \to \R^\dimInd$ be any continuous mappings and let $\density : \R^\dimInd\to \R$ be any probability density.

Then,
    \begin{equation}
    W_1((\RAEdecoder \circ \RAEencoder)_{\sharp} \density, \density) \leq \mathbb{E}_{\stoVector\sim \density}\Bigl[ \|\RAEdecoder(\RAEencoder(\stoVector)) - \stoVector\|_2\Bigr]
    \label{eq:W1-bound}
\end{equation}
\end{proposition}

\begin{proof}
    Choosing the coupling
    \begin{equation}
        \gamma_0(\Vector, \Vector') := \density(\Vector) \delta(\RAEdecoder(\RAEencoder(\Vector)) - \Vector')
    \end{equation}
    we have
    \begin{multline}
        W_1((\RAEdecoder \circ \RAEencoder)_{\sharp} \density, \density) \leq \mathbb{E}_{(\stoVector, \stoVector')\sim \gamma_0} [\|\stoVector - \stoVector'\|_2] \\
        = \int_{\R^\dimInd\times \R^\dimInd} \density(\Vector) \delta(\RAEdecoder(\RAEencoder(\Vector)) - \Vector') \|\Vector - \Vector'\|_2 \; \mathrm{d}\Vector \mathrm{d}\Vector' \\
        = \int_{\R^\dimInd} \density(\Vector) \|\RAEdecoder(\RAEencoder(\Vector)) - \Vector\|_2 \; \mathrm{d}\Vector 
        = \mathbb{E}_{\stoVector\sim \density}\Bigl[ \|\RAEdecoder(\RAEencoder(\stoVector)) - \stoVector\|_2\Bigr],
    \end{multline}
    which proves the claim.
\end{proof}

The above results tells us that constructing a suitable RAE boils down to finding a suitable base point and orthonormal basis that -- when combined with our diffeomorphism -- yield a small expected projection error. In addition, we aim to bound this expected projection error in terms of $\spdMatrix$ and $\diffeo$ -- and without explicitly invoking the latent space dimension. To achieve this, we will start by evaluating the tangent space covariant matrix at a special choice of base point. 

\begin{lemma}[Tangent space covariance matrix]
\label{lem:cov-matrix-on-data-space}
    Let $\density:\R^\dimInd \to \R$ be a probability density of the form \cref{eq:paper-density-model} -- generated by a smooth diffeomorphism $\diffeo:\R^\dimInd \to \R^\dimInd$ and positive definite matrix $\spdMatrix \in \R^{\dimInd\times \dimInd}$ -- and define $\bar{\Vector} = \diffeo^{-1}(\mathbf{0})$.

    Then, the covariance matrix at the tangent space $\tangent_{\bar{\Vector}} \R^{\dimInd}$ satisfies
    \begin{equation}
        \mathbb{E}_{\stoVector \sim \density} \Bigl[ \log_{\bar{\Vector}}^\diffeo (\stoVector) \otimes \log_{\bar{\Vector}}^\diffeo (\stoVector) \Bigr] = D_{\mathbf{0}} \diffeo^{-1} \spdMatrix (D_{\mathbf{0}} \diffeo^{-1})^\top.
    \end{equation}
\end{lemma}

\begin{proof}
    The proof follows from direct computation: 
    \begin{multline}
        \mathbb{E}_{\stoVector \sim \density} \Bigl[ \log_{\bar{\Vector}}^\diffeo (\stoVector) \otimes \log_{\bar{\Vector}}^\diffeo (\stoVector) \Bigr] 
        = \mathbb{E}_{\stoVector \sim \density} \Bigl[D_{\mathbf{0}}\diffeo^{-1}[\diffeo(\stoVector)] \otimes D_{\mathbf{0}}\diffeo^{-1}[\diffeo(\stoVector)]\Bigr]\\
        = \mathbb{E}_{\stoLatentVector \sim \mathcal{N}(0, \spdMatrix)} \Bigl[D_{\mathbf{0}}\diffeo^{-1}[\stoLatentVector] \otimes D_{\mathbf{0}}\diffeo^{-1}[\stoLatentVector]\Bigr] 
        = D_{\mathbf{0}}\diffeo^{-1} \mathbb{E}_{\stoLatentVector \sim \mathcal{N}(0, \spdMatrix)} [\stoLatentVector \otimes \stoLatentVector](D_{\mathbf{0}}\diffeo^{-1})^\top
        \\
        = D_{\mathbf{0}} \diffeo^{-1} \spdMatrix (D_{\mathbf{0}} \diffeo^{-1})^\top.
    \end{multline}
\end{proof}



    



Using this covariance matrix -- which tells us where the data live when looking from $\bar{\Vector}$ -- we follow a RAE construction similar to \cite{diepeveen2025scorebased}. In particular, we consider the eigendecomposition  $D_{\mathbf{0}} \diffeo^{-1}\spdMatrix(D_{\mathbf{0}} \diffeo^{-1})^\top = \mathbf{U}\spdMatrixB\mathbf{U}^\top$ with
\begin{equation}
    \spdMatrixB_{1, 1}\geq \ldots \geq \spdMatrixB_{\dimInd, \dimInd}> 0,
\end{equation}
to define a family of RAEs. That is, for any $\RAErelerror\in (\RAErelerror_0,1]$ with $\RAErelerror_0 := \frac{\spdMatrixB_{\dimInd, \dimInd} }{\operatorname{tr} (\spdMatrixB)}$ we define $\dimInd_\RAErelerror \in [\dimInd-1]$ as the integer that satisfies
\begin{equation}
    \dimInd_\RAErelerror := 
 \min \Bigl\{ \dimInd'\in [\dimInd-1] \; \Bigl\vert \; \sum_{\sumIndC=\dimInd' + 1}^{\dimInd} \spdMatrixB_{\sumIndC, \sumIndC}  \leq \RAErelerror \operatorname{tr} (\spdMatrixB) \Bigr\},
\label{eq:dimind-epsilon}
\end{equation}
and define the mappings $\RAEencoder_\RAErelerror:\R^\dimInd \to \R^{\dimInd_\RAErelerror}$ and $\RAEdecoder_\RAErelerror:\R^{\dimInd_\RAErelerror} \to \R^\dimInd$ as
\begin{equation}
    \RAEencoder_\RAErelerror(\Vector) := \mathbf{U}_\RAErelerror^\top \log^{\diffeo}_{\bar{\Vector}} (\Vector) \quad \text{and} \quad  \RAEdecoder_\RAErelerror(\latentVector):= \exp_{\bar{\Vector}}^\diffeo \Bigl( \mathbf{U}_\RAErelerror \latentVector\Bigr),
    \label{eq:rae-encoder-nf}
\end{equation}
where $\mathbf{U}_\RAErelerror := [\mathbf{u}^{1}, \ldots, \mathbf{u}^{\dimInd_\RAErelerror}] \in(\tangent_{\bar{\Vector}} \R^\dimInd)^{\dimInd_\RAErelerror} \cong \R^{\dimInd \times \dimInd_\RAErelerror}$, and which together generate a family of Riemannian Autoencoders parametrized by $\RAErelerror$.

The choice of $\RAErelerror$ does not only control the dimension of the latent space of the RAE, but also governs the expected projection error, along with $\spdMatrix$ and $\diffeo$. To see this, we will first prove the following auxiliary lemma.




\begin{lemma}[Expected tangent space projection error]
\label{lem:thm-expected-rae-error-step3}
Let $\density:\R^\dimInd \to \R$ be any probability density of the form \cref{eq:paper-density-model} -- generated by a smooth diffeomorphism $\diffeo:\R^\dimInd \to \R^\dimInd$ and positive definite matrix $\spdMatrix \in \R^{\dimInd\times \dimInd}$ -- and define $\bar{\Vector} = \diffeo^{-1}(\mathbf{0})$. Furthermore, consider any $\RAErelerror \in (\RAErelerror_0,1]$ and define the mappings $\RAEencoder_\RAErelerror:\R^\dimInd \to \R^{\dimInd_\RAErelerror}$ and $\RAEdecoder_\RAErelerror:\R^{\dimInd_\RAErelerror} \to \R^\dimInd$ as in \cref{eq:rae-encoder-nf} for $\dimInd_\RAErelerror$ as in \cref{eq:dimind-epsilon}.

Then,
    \begin{equation}
    \mathbb{E}_{\stoVector\sim \density} \Bigl[ \|\sum_{\sumIndC=\dimInd_\RAErelerror+1}^{\dimInd} (\log^\diffeo_{\bar{\Vector}} (\stoVector), \mathbf{u}^{\sumIndC})_2 \mathbf{u}^{\sumIndC}]\|_2^2 \Bigr] 
     \leq \RAErelerror  \|D_{\mathbf{0}}\diffeo^{-1} \spdMatrix^{\frac{1}{2}}\|_F^2.
    \end{equation}
\end{lemma}

\begin{proof} We have
    
\begin{multline}
    \mathbb{E}_{\stoVector\sim \density}\Bigl[ \|\sum_{\sumIndC=\dimInd_\RAErelerror+1}^{\dimInd} (\log^\diffeo_{\bar{\Vector}} (\stoVector), \mathbf{u}^{\sumIndC})_2 \mathbf{u}^{\sumIndC}]\|_2^2 \Bigr] 
    = \mathbb{E}_{\stoVector\sim \density}\Bigl[ \sum_{\sumIndC=\dimInd_\RAErelerror+1}^{\dimInd} (\log^\diffeo_{\bar{\Vector}} (\stoVector), \mathbf{u}^{\sumIndC})_2^2 \Bigr] \qquad \qquad \qquad \qquad\\
    = \sum_{\sumIndC=\dimInd_\RAErelerror+1}^{\dimInd} (\mathbf{u}^{\sumIndC})^\top\mathbb{E}_{\stoVector\sim \density}\Bigl[  \log^\diffeo_{\bar{\Vector}} (\stoVector) \otimes \log^\diffeo_{\bar{\Vector}} (\stoVector) \Bigr] \mathbf{u}^{\sumIndC}
    \overset{\text{\cref{lem:cov-matrix-on-data-space}}}{=} \sum_{\sumIndC=\dimInd_\RAErelerror+1}^{\dimInd} (\mathbf{u}^{\sumIndC})^\top D_{\mathbf{0}} \diffeo^{-1} \spdMatrix (D_{\mathbf{0}} \diffeo^{-1})^\top \mathbf{u}^{\sumIndC}\\
    = \sum_{\sumIndC=\dimInd_\RAErelerror+1}^{\dimInd} \spdMatrixB_{\sumIndC, \sumIndC} \leq \RAErelerror \operatorname{tr}(\spdMatrixB)
    = \RAErelerror \operatorname{tr}(D_{\mathbf{0}} \diffeo^{-1} \spdMatrix (D_{\mathbf{0}} \diffeo^{-1})^\top)
    = \RAErelerror  \|D_{\mathbf{0}}\diffeo^{-1} \spdMatrix^{\frac{1}{2}}\|_F^2
\end{multline} where the inequality follows by definition of $d_{\varepsilon}$.
\end{proof}

Finally, we are ready to show that the expected projection error is indeed governed by $\RAErelerror$, $\spdMatrix$ and $\diffeo$. In particular, the following result tells us that for a given $\RAErelerror$, the corresponding RAE will have a low expected projection error if: (i) the diffeomorphism $\diffeo$ is very smooth, and (ii) the matrix $D_{\mathbf{0}}\diffeo^{-1} \spdMatrix^{\frac{1}{2}}$ is approximately low-rank.

\begin{theorem}[Expected projection error] 
\label{thm:expected-proj-err}
Let $\density:\R^\dimInd \to \R$ be a probability density of the form \cref{eq:paper-density-model} -- generated by a smooth diffeomorphism $\diffeo:\R^\dimInd \to \R^\dimInd$ and positive definite matrix $\spdMatrix \in \R^{\dimInd\times \dimInd}$ -- and define $\bar{\Vector} = \diffeo^{-1}(\mathbf{0})$. Furthermore, consider any $\RAErelerror \in (\RAErelerror_0,1]$ and define the mappings $\RAEencoder_\RAErelerror:\R^\dimInd \to \R^{\dimInd_\RAErelerror}$ and $\RAEdecoder_\RAErelerror:\R^{\dimInd_\RAErelerror} \to \R^\dimInd$ as in \cref{eq:rae-encoder-nf} for $\dimInd_\RAErelerror$ as in \cref{eq:dimind-epsilon}.

Then,
    \begin{multline}
    \mathbb{E}_{\stoVector\sim \density}\Bigl[ \|\RAEdecoder_\RAErelerror(\RAEencoder_\RAErelerror(\stoVector)) - \stoVector\|_2\Bigr] \leq C_{\RAErelerror}^{(1)} \sqrt{\RAErelerror}\|D_{\mathbf{0}}\diffeo^{-1} \spdMatrix^{\frac{1}{2}}\|_F 
    + \frac{1}{2}C_{\RAErelerror}^{(2)} \RAErelerror \|D_{\mathbf{0}}\diffeo^{-1} \spdMatrix^{\frac{1}{2}}\|_F^2
    + o(\RAErelerror),
\end{multline}
where 
\begin{align}
    C_{\RAErelerror}^{(1)} &= \sup_{\latentVector\in \R^{\dimInd_\RAErelerror}}\|D_{\diffeo(\RAEdecoder_\RAErelerror(\latentVector))}\diffeo^{-1} [D_{\bar{\Vector}}\diffeo[\cdot]]\|_2, \\
    C_{\RAErelerror}^{(2)} &= \sup_{\latentVector\in \R^{\dimInd_\RAErelerror}}\|D^2_{\diffeo(\RAEdecoder_\RAErelerror(\latentVector))}\diffeo^{-1} [D_{\bar{\Vector}}\diffeo[\cdot], D_{\bar{\Vector}}\diffeo[\cdot]]\|_2.
\end{align}
\end{theorem}

\begin{proof}
    First, for a given $\Vector$, we define $\dummyVector: = \diffeo(\RAEdecoder_\RAErelerror(\RAEencoder_\RAErelerror(\Vector)))$ and $\dummyVectorB := \diffeo(\Vector) - \dummyVector$ and note that we have 
    \begin{equation}
        \dummyVector= D_{\bar{\Vector}}\diffeo[\sum_{\sumIndC=1}^{\dimInd_\RAErelerror} (\log^\diffeo_{\bar{\Vector}} (\Vector), \mathbf{u}^{\sumIndC})_2 \mathbf{u}^{\sumIndC}]
    \end{equation}
    and 
    \begin{equation}
        \dummyVectorB = D_{\bar{\Vector}}\diffeo[\sum_{\sumIndC=\dimInd_\RAErelerror+1}^{\dimInd} (\log^\diffeo_{\bar{\Vector}} (\Vector), \mathbf{u}^{\sumIndC})_2 \mathbf{u}^{\sumIndC}].
    \end{equation}
    Next, we can rewrite
    \begin{equation}
        \|\RAEdecoder_\RAErelerror(\RAEencoder_\RAErelerror(\Vector)) - \Vector\|_2 = \|\diffeo^{-1} (\dummyVector) - \diffeo^{-1} (\dummyVector + \dummyVectorB)\|_2,
        \label{thm-expected-rae-error-step1}
    \end{equation}
and note that a Taylor expansion of $\diffeo^{-1}$ at $\dummyVector$ gives
\begin{equation}
     \diffeo^{-1} (\dummyVector + \dummyVectorB) - \diffeo^{-1} (\dummyVector) = D_{\dummyVector} \diffeo^{-1} [\dummyVectorB] + \frac{1}{2} D_{\dummyVector}^2 \diffeo^{-1} [\dummyVectorB, \dummyVectorB]
     + \mathcal{O} (\|\dummyVectorB\|_2^3).
\end{equation}


Then, by triangle inequality we see that
\begin{equation}
    \|\diffeo^{-1} (\dummyVector) - \diffeo^{-1} (\dummyVector + \dummyVectorB)\|_2 \leq \|D_{\dummyVector} \diffeo^{-1} [\dummyVectorB]\|_2 
    + \frac{1}{2} \|D_{\dummyVector}^2 \diffeo^{-1} [\dummyVectorB, \dummyVectorB]\|_2
     + \mathcal{O} (\|\dummyVectorB\|_2^3).
      \label{thm-expected-rae-error-step2}
\end{equation}
In addition, we have
\begin{multline}
    \|D_{\dummyVector} \diffeo^{-1} [\dummyVectorB]\|_2 \leq  \|D_{\dummyVector} \diffeo^{-1} D_{\bar{\Vector}}\diffeo \|_2  \|\sum_{\sumIndC=\dimInd_\RAErelerror+1}^{\dimInd} (\log^\diffeo_{\bar{\Vector}} (\Vector), \mathbf{u}^{\sumIndC})_2 \mathbf{u}^{\sumIndC}\|_2\\
     \leq C_{\RAErelerror}^{(1)} \|\sum_{\sumIndC=\dimInd_\RAErelerror+1}^{\dimInd} (\log^\diffeo_{\bar{\Vector}} (\Vector), \mathbf{u}^{\sumIndC})_2 \mathbf{u}^{\sumIndC}\|_2,
     \label{thm-expected-rae-error-step2a}
\end{multline}
and
\begin{multline}
    \|D_{\dummyVector}^2 \diffeo^{-1} [\dummyVectorB, \dummyVectorB]\|_2 \leq \|D_{\dummyVector}^2 \diffeo^{-1}[D_{\bar{\Vector}}\diffeo [\cdot],D_{\bar{\Vector}}\diffeo [\cdot]] \|_2 \|\sum_{\sumIndC=\dimInd_\RAErelerror+1}^{\dimInd} (\log^\diffeo_{\bar{\Vector}} (\Vector), \mathbf{u}^{\sumIndC})_2 \mathbf{u}^{\sumIndC}\|_2^2\\
    \leq C_{\RAErelerror}^{(2)} \|\sum_{\sumIndC=\dimInd_\RAErelerror+1}^{\dimInd} (\log^\diffeo_{\bar{\Vector}} (\Vector), \mathbf{u}^{\sumIndC})_2 \mathbf{u}^{\sumIndC}\|_2^2.
    \label{thm-expected-rae-error-step2b}
\end{multline}

Finally, combining the above results, we have that
\begin{multline}
    \mathbb{E}_{\stoVector\sim \density}\Bigl[ \|\RAEdecoder_\RAErelerror(\RAEencoder_\RAErelerror(\stoVector)) - \stoVector\|_2\Bigr] \\
    \overset{\text{\cref{thm-expected-rae-error-step1}-\cref{thm-expected-rae-error-step2b}}}{\leq} \mathbb{E}_{\stoVector\sim \density}\Bigl[ C_{\RAErelerror}^{(1)} \|\sum_{\sumIndC=\dimInd_\RAErelerror+1}^{\dimInd} (\log^\diffeo_{\bar{\Vector}} (\stoVector), \mathbf{u}^{\sumIndC})_2 \mathbf{u}^{\sumIndC}\|_2 
 + \frac{1}{2}C_{\RAErelerror}^{(2)} \|\sum_{\sumIndC=\dimInd_\RAErelerror+1}^{\dimInd} (\log^\diffeo_{\bar{\Vector}} (\stoVector), \mathbf{u}^{\sumIndC})_2 \mathbf{u}^{\sumIndC}\|_2^2 \\
 + \mathcal{O}(\|\sum_{\sumIndC=\dimInd_\RAErelerror+1}^{\dimInd} (\log^\diffeo_{\bar{\Vector}} (\stoVector), \mathbf{u}^{\sumIndC})_2 \mathbf{u}^{\sumIndC}]\|_2^3) \Bigr]\\
 \overset{\text{\cref{lem:thm-expected-rae-error-step3} and Cauchy-Schwarz}}{\leq} C_{\RAErelerror}^{(1)} \sqrt{\RAErelerror}\|D_{\mathbf{0}}\diffeo^{-1} \spdMatrix^{\frac{1}{2}}\|_F
    + \frac{1}{2}C_{\RAErelerror}^{(2)} \RAErelerror \|D_{\mathbf{0}}\diffeo^{-1} \spdMatrix^{\frac{1}{2}}\|_F^2 + o(\RAErelerror),
\end{multline}
which proves the claim.

\end{proof}

\subsection{Joint Manifold Learning and Generative Modeling}

Now that we have a systematic way of constructing Riemannian Autoencoders (RAE) given a probability density of the form \cref{eq:paper-density-model} and are able to control its expected projection error through $\spdMatrix$ and $\diffeo$ -- assuming a fixed $\RAErelerror$ --, our next step is to integrate these ideas into the AmbientFlow \cite{kelkarambientflow} framework. In particular, in the following we will consider the optimization problem
\begin{align}
    \inf_{\networkParams,\networkParamsB, \networkParamsC} & -\mathcal{L}_{\text{VLB}}(\networkParams,\networkParamsB) \nonumber \\ 
    \text{s.t.} & \quad \mathbb{E}_{\stoVector \sim \density_{\networkParams}}\Bigl[\|\RAEdecoder_{\RAErelerror,\networkParamsC}(\RAEencoder_{\RAErelerror,\networkParamsC}(\stoVector)) - \stoVector\|_2 
    \label{eq:rie-ambientflow-problem} \\
    & \qquad + \|(\diffeo_{\networkParamsC} \circ \diffeo^{-1}_{\networkParams}) (\stoVector) - \stoVector\|_2\Bigr] \leq \omega, \nonumber
\end{align} 
where the objective function $\mathcal{L}_{\text{VLB}}$ is the variational lower bound \cref{eq:vlb-ambientflow} -- underlying AmbientFlow -- that jointly couples the learnable prior density $\density_{\networkParams}:\R^\dimInd\to\R$ -- which is of the form \cref{eq:paper-density-model} with unit positive definite matrix $\mathbf{I} \in \R^{\dimInd\times \dimInd}$ and learnable diffeomorphic neural network $\diffeo_{\networkParams}: \R^\dimInd\to\R^\dimInd$ -- and the learnable posterior density $\density_{\networkParamsB} (\cdot\,|\,\cdot): \R^\dimInd\times \R^{\dimIndB} \to \R$. The main difference in our framework is the third density $\density_\networkParamsC:\R^\dimInd\to\R$ -- which is also of the form \cref{eq:paper-density-model} with unit positive definite matrix $\mathbf{I} \in \R^{\dimInd\times \dimInd}$, but has a learnable diffeomorphic neural network $\diffeo_{\networkParamsC}: \R^\dimInd\to\R^\dimInd$ -- and is used to parametrize the RAE encoder and decoder\footnote{The additional subscript $\networkParamsC$ in the encoder and decoder are there to indicate that they are constructed through $\density_\networkParamsC$.}. Finally, $\omega >0$ controls the level of geometric regularization enforced on the learned density $p_{\networkParams}$.


To generalize the recoverability result under a sparsity constraint \cite[Thm.~3.2]{kelkarambientflow}, we will need a generalization of the Restricted Isometry Property (RIP) that fits more naturally within our Riemannian framework.

\begin{definition}[Restricted Isometry Property] 
    Let $\RAEdecoder:\R^{\dimIndC} \to \R^\dimInd$ be any continuous mapping. A matrix $\forward\in \R^{\dimIndB\times\dimInd}$ satisfies the \emph{restricted isometry property with constant $\delta \in (0,1)$ with respect to the range of $\RAEdecoder$} if for all $\Vector_1,\Vector_2 \in \operatorname{range}(\RAEdecoder)$
    \begin{equation}
        (1 - \delta)\|\Vector_1 - \Vector_2\|_2^2 \leq \|\forward\Vector_1 - \forward\Vector_2\|_2^2 \leq (1 + \delta)\|\Vector_1 - \Vector_2\|_2^2.
    \end{equation}
    
\end{definition} 
For generic, random measurements, this condition often holds when the number of measurements scales with the (squared) Gaussian width of the signal set $m \geq \Omega(\omega^2(\mathrm{range}(\RAEdecoder_\RAErelerror))$. For our smooth, Lipschitz decoder $\RAEdecoder_\RAErelerror$, $\omega^2(\mathrm{range}(\RAEdecoder_\RAErelerror)) \propto \latentdim \log(\mathrm{Lip}(\RAEdecoder_\RAErelerror))$ (see, e.g., \cite{bora2017compressed}) as it is a smooth Lipschitz map (to be discussed in \cref{sec:decoder-properties}).

Under this RIP assumption, the following result shows that, under reasonable technical conditions -- discussed in more detail after the proof -- the method recovers essentially the correct data distribution from indirect, noisy measurements, with a controllable error that grows at most linearly in the tolerance parameter $\omega$.

\begin{theorem}[Recoverability]
\label{thm:recoverability}
    Consider any minimizer $(\hat{\networkParams},\hat{\networkParamsB},\hat{\networkParamsC})$ to the problem \cref{eq:rie-ambientflow-problem} and
    assume the following hold:
    \begin{itemize}
        \item $\forward$ satisfies RIP with constant $\delta \in (0,1)$ with respect to $\mathrm{range}(\RAEdecoder_{\RAErelerror,\hat{\networkParamsC}})$.
        \item There exist a parametrization $(\networkParams_*, \networkParamsB_*)$ such that both $\density_{\networkParams_*} = \density_{\text{data}}$ and $\density_{\networkParamsB_*}(\, \cdot\,|\,\stoCorVector) = \density_{\networkParams_*}(\, \cdot\,|\,\stoCorVector)$, and such that $(\networkParams_*,\networkParamsB_*,\hat{\networkParamsC})$ is feasible.
        \item The characteristic function of the noise distribution $\chi_{\mathbf{n}}$ has full support over $\R^\dimIndB$.
    \end{itemize} Then, the distribution $\density_{\hat{\networkParams}}$ recovered by $\hat{\networkParams}$ satisfies 
    \begin{equation}
        W_1(\density_{\hat{\networkParams}}, \density_{\text{data}} ) \leq 2 \omega \left(1 + \frac{\|\forward\|}{\sqrt{1-\delta}}\right).
        \label{eq:thm-recov-main-result}
    \end{equation}
\end{theorem}
\begin{proof}
First, since $(\networkParams_*, \networkParamsB_*)$ satisfying $p_{\networkParams_*}=\density_{\text{data}}$, and $\density_{\networkParamsB_*}(\, \cdot\,|\,\stoCorVector) = \density_{\networkParams_*}(\, \cdot\,|\,\stoCorVector)$ is a feasible solution to \cref{eq:rie-ambientflow-problem}, the maximum value of $\mathcal{L}_{\text{VLB}}$ under \cref{eq:rie-ambientflow-problem} is $\mathbb{E}_{\stoCorVector \sim \densityB_{\text{data}}} \log \densityB_{\text{data}}(\stoCorVector)$. Therefore, for the estimated $\hat{\networkParams}$ and $\hat{\networkParamsB}$, we must also have  that $\mathcal{L}_{\text{VLB}}(\hat{\theta}, \hat{\phi})=\mathbb{E}_{\stoCorVector \sim \densityB_{\text{data}}} \log \densityB_{\text{data}}(\stoCorVector)$ and that
\begin{equation}
    \forward_{\sharp}(\density_{\hat{\networkParams}}) * \density_{\text{noise}}=\densityB_{\text{data}} \text { and } p_{\hat{\networkParamsB}}(\cdot \mid \CorVector)=p_{\hat{\networkParams}}(\cdot \mid \CorVector) .
    \label{eq:cor-densities-equal}
\end{equation}

Then, to show our claim \cref{eq:thm-recov-main-result}, we will for any $\Vector_1,\Vector_2 \in \mathbb{R}^\dimInd$ write $\hat{\Vector}_i:= \RAEdecoder_{\RAErelerror,\hat{\networkParamsC}}(\RAEencoder_{\RAErelerror,\hat{\networkParamsC}}(\Vector_i))$, $\CorVector_i := \forward\Vector_i$, and $\hat{\CorVector}_i = \forward\hat{\Vector}_i$, and consider the bound 
 $$
 \|\Vector_1 - \Vector_2\|_2 \leq \|\Vector_1 - \hat{\Vector}_1\|_2 + \|\Vector_2 - \hat{\Vector}_2\|_2 + \|\hat{\Vector}_1 - \hat{\Vector}_2\|_2.
 $$ 
 Next, consider a joint distribution $\gamma \in \Gamma(\density_{\hat{\networkParams}}, \density_{\text{data}})$. We have  
 \begin{multline}
        \mathbb{E}_{(\stoVector_1,\stoVector_2)\sim \gamma}\|\stoVector_1 - \stoVector_2\|_2= \mathbb{E}_{\stoVector_1 \sim \density_{\text{data}}}\|\stoVector_1 - \hat{\stoVector}_1\|_2
        + \mathbb{E}_{\stoVector_2 \sim \density_{\hat{\networkParams}}}\|\stoVector_2 - \hat{\stoVector}_2\|_2\\
        + \mathbb{E}_{(\stoVector_1,\stoVector_2)\sim\gamma}\|\hat{\stoVector}_1 - \hat{\stoVector}_2\|_2.
    \end{multline} 
    For the first term
    \begin{multline}
      \mathbb{E}_{\stoVector_1\sim \density_{\text{data}}}\|\stoVector_1 - \hat{\stoVector}_1\|_2  \leq \mathbb{E}_{\stoVector_1\sim \density_{\text{data}}}\|\stoVector_1 - \RAEdecoder_{\RAErelerror,\hat{\networkParamsC}}(\RAEencoder_{\RAErelerror,\hat{\networkParamsC}} (\stoVector_1))\|_2 \\
        = \mathbb{E}_{\Vector_1 \sim \density_{\networkParams_*}}\|\stoVector_1 - \RAEdecoder_{\RAErelerror,\hat{\networkParamsC}}(\RAEencoder_{\RAErelerror,\hat{\networkParamsC}}(\stoVector_1))\|_2  \leq \omega,
        \label{eq:thm-distribution-retrieval-error-step1}
    \end{multline} 
     since $\density_{\networkParams_*} = \density_{\text{data}}$ is feasible.
    For the second term we directly get
    \begin{equation}
        \mathbb{E}_{\stoVector_2\sim \density_{\hat{\networkParams}}}\|\stoVector_2 - \RAEdecoder_{\RAErelerror,\hat{\networkParamsC}}(\RAEencoder_{\RAErelerror,\hat{\networkParamsC}}(\stoVector_2))\|_2 \leq \omega.
        \label{eq:thm-distribution-retrieval-error-step2}
    \end{equation} 
    By our RIP assumption on $\forward$, since $\hat{\Vector}_i \in \mathrm{range}(\RAEdecoder_{\RAErelerror,\hat{\networkParamsC}})$, note that we have 
    \begin{equation}
        \|\hat{\Vector}_1 - \hat{\Vector}_2\|_2\leq \frac{1}{\sqrt{1-\delta}}\|\hat{\CorVector}_1 - \hat{\CorVector}_2\|_2.
        \label{eq:thm-distribution-retrieval-error-step3a}
    \end{equation}
    Then 
    \begin{multline}
        \|\hat{\CorVector}_1 - \hat{\CorVector}_2\|_2 \leq \|\CorVector_1 - \hat{\CorVector}_1\|_2+ \|\CorVector_2 - \hat{\CorVector}_2\|_2+ \|\CorVector_1 - \CorVector_2\|_2 \\
         \leq \|\forward\|\cdot(\|\Vector_1 - \hat{\Vector}_1\|_2+ \|\Vector_2 - \hat{\Vector}_2\|) 
        + \|\CorVector_1 - \CorVector_2\|_2.
    \end{multline} 
    Taking an expectation over $\gamma$ and using our two previous results gives 
    \begin{equation}
        \mathbb{E}_{(\stoVector_1,\stoVector_2)\sim\gamma}\|\hat{\stoCorVector}_1 - \hat{\stoCorVector}_2\|_2\overset{\cref{eq:thm-distribution-retrieval-error-step1} + \cref{eq:thm-distribution-retrieval-error-step2}}{\leq} 2\|\forward\|\omega + \mathbb{E}_{(\stoVector_1,\stoVector_2)\sim\gamma}\|\stoCorVector_1 - \stoCorVector_2\|_2, 
        \label{eq:thm-distribution-retrieval-error-step3b}
    \end{equation} 
    from which we find that 
    \begin{equation}
        \mathbb{E}_{(\stoVector_1,\stoVector_2)\sim\gamma} \|\hat{\stoVector}_1-\hat{\stoVector}_2\|_2\overset{\cref{eq:thm-distribution-retrieval-error-step3a} + \cref{eq:thm-distribution-retrieval-error-step3b}}{\leq} \frac{2\|\forward\|}{\sqrt{1-\delta}} \omega 
        + \frac{1}{\sqrt{1-\delta}}\mathbb{E}_{(\stoVector_1,\stoVector_2)\sim\gamma}\|\stoCorVector_1 - \stoCorVector_2\|_2.
        \label{eq:thm-distribution-retrieval-error-step3}
    \end{equation}
    Combining our results and taking an infimum over $\gamma \in \Gamma(\density_{\hat{\networkParams}},\density_{\text{data}})$ gives \begin{multline}
        W_1(\density_{\hat{\networkParams}}, \density_{\text{data}})  = \inf_{\gamma \in \Gamma(\density_{\hat{\networkParams}},\density_{\text{data}})}\mathbb{E}_{(\stoVector_1,\stoVector_2)\sim\gamma}\|\stoVector_1-\stoVector_2\|_2\\
         \leq \mathbb{E}_{\stoVector\sim\density_{\text{data}}}\|\stoVector_1 - \hat{\stoVector}_1\|_2+ \mathbb{E}_{\stoVector\sim\density_{\hat{\networkParams}}}\|\stoVector_2 - \hat{\stoVector}_2\|_2
         + \inf_{\gamma \in \Gamma(\density_{\hat{\networkParams}},\density_{\text{data}})}\mathbb{E}_{(\stoVector_1,\stoVector_2) \sim \gamma}\|\hat{\stoVector}_1 - \hat{\stoVector}_2\|_2\\
         \overset{\cref{eq:thm-distribution-retrieval-error-step1} + \cref{eq:thm-distribution-retrieval-error-step2} + \cref{eq:thm-distribution-retrieval-error-step3}}{\leq} 2 \omega \left(1 + \frac{\|\forward\|}{\sqrt{1-\delta}}\right) 
         + \frac{1}{\sqrt{1-\delta}}\inf_{\gamma \in \Gamma(\density_{\hat{\networkParams}},\density_{\text{data}})}\mathbb{E}_{(\stoVector_1,\stoVector_2) \sim\gamma}\|\forward\stoVector_1 - \forward\stoVector_2\|_2.
    \end{multline} 
    The final term can be shown to be zero using \cref{eq:cor-densities-equal} and the final assumption on full support of the noise -- as in the end of the proof of \cite[Thm.~1]{kelkarambientflow}. This gives 
    \begin{equation}
        W_1(\density_{\hat{\networkParams}}, \density_{\text{data}}) \leq 2 \omega \left(1 + \frac{\|\forward\|}{\sqrt{1-\delta}}\right) .
    \end{equation}
\end{proof}

While this theorem provides a reassuring theoretical guarantee, several caveats are worth noting. First, problem \cref{eq:rie-ambientflow-problem} is not particularly practical to implement, as it relies on RAEs whose latent dimensions vary with $\diffeo_\networkParamsC$. Second, the subtlety in \cref{thm:recoverability} lies in its second assumption: it effectively says that if $\diffeo_\networkParamsC$ happens to define an RAE that can compress the ground-truth distribution, then the guarantee holds, but the optimization problem itself does not ensure that this condition is met. Third, in practice, the data distribution may also deviate from the idealized model class in \cref{eq:paper-density-model}, potentially degrading the quality of the RAE construction, which critically relies on this assumption.



\paragraph{Practical Implementation} In practice, the constraint in \cref{eq:rie-ambientflow-problem} is not particularly tractable, even though it greatly simplifies the theoretical analysis. However, parametrizing $\diffeo_{\networkParamsC} := \diffeo_{\networkParams}$ we get
\begin{multline}
    \mathbb{E}_{\stoVector \sim \density_{\networkParams}(\cdot)}\left[\|\RAEdecoder_{\RAErelerror,\networkParamsC}(\RAEencoder_{\RAErelerror,\networkParamsC}(\stoVector)) - \stoVector\|_2 + \|(\diffeo_{\networkParamsC} \circ \diffeo^{-1}_{\networkParams}) (\stoVector) - \stoVector\|_2\right]\\
    = \mathbb{E}_{\stoVector \sim \density_{\networkParams}(\cdot)}\left[\|\RAEdecoder_{\RAErelerror,\networkParams}(\RAEencoder_{\RAErelerror,\networkParams}(\stoVector)) - \stoVector\|_2 \right] 
\end{multline}
which we know from \cref{thm:expected-proj-err} is governed by $\|D_{\mathbf{0}}\diffeo_\networkParams^{-1} \spdMatrix^{\frac{1}{2}}\|_F$ for small enough $\RAErelerror$, regular enough $\diffeo_\networkParams$, and low-rank $D_{\mathbf{0}}\diffeo_\networkParams^{-1} \spdMatrix^{\frac{1}{2}}$.
Using that we assume $\spdMatrix = \mathbf{I}$ we can rewrite the optimization problem \cref{eq:rie-ambientflow-problem} as
\begin{align}
     \inf_{\networkParams,\networkParamsB} & -\mathcal{L}_{\text{VLB}}(\networkParams,\networkParamsB) 
        \\
        \text{s.t.} & \quad \sqrt{\RAErelerror} \|D_{\mathbf{0}}\diffeo_\networkParams^{-1}\|_F \leq \omega'\nonumber
    \end{align} 
    or equivalently
    \begin{equation}
        \inf_{\networkParams,\networkParamsB} - \mathcal{L}_{\text{VLB}}(\networkParams,\networkParamsB) + \lambda \|D_{\mathbf{0}}\diffeo_\networkParams^{-1}\|_F
        \label{eq:updated-rie-ambient-flow}
    \end{equation}
    for some $\lambda>0$, which is much more tractable.\

    \begin{remark}
        For this line of reasoning to yield a valid solution, it is crucial to work with regular parametrizations for $\diffeo_\networkParams$, in accordance with earlier observations on learning pullback geometries \cite{diepeveen2024pulling} and with current practice \cite{diepeveen2025scorebased,diepeveen2025manifold}.
    \end{remark}

    \paragraph{Availability of Limited Reference Data}
    We should expect that the second caveat still holds in the new setting \cref{eq:updated-rie-ambient-flow} since we are still not sure which local minimum we end up in, let alone whether the assumptions on feasibility are satisfied. In order to alleviate this, we can easily add extra regularization if we have access to limited reference data, e.g., simulated data or real ground truth data, through adding a standard negative log likelihood term. This has been commonly exploited in several recent works on learning generative models from corrupted data \cite{weiminbai2024emdiffusion, daras2025ambientscaling, lustochastic}. Then, the full Riemannian AmbientFlow loss is given by
    \begin{equation}
        \inf_{\networkParams,\networkParamsB} - \mathcal{L}_{\text{VLB}}(\networkParams,\networkParamsB) + \lambda \|D_{\mathbf{0}}\diffeo_\networkParams^{-1}\|_F
        - \mu \mathbb{E}_{\stoVector \sim \density_{\text{ref}}} [\log (\density_\networkParams (\stoVector))],
        \label{eq:full-rie-ambientflow-loss}
    \end{equation}
    for some $\mu >0$. 

    \paragraph{Constructing an RAE after Training}
    In principle, one could construct an RAE directly from the assumed model -- as outlined above --, but in practice the data distribution rarely matches this form exactly, which leads to a suboptimal RAE. To overcome this, data are needed to fit the RAE, and these can be obtained from: (i) the limited reference data, or (ii) the posterior density $\density_{\networkParamsB} (\cdot\,|\,\cdot): \R^\dimInd\times \R^{\dimIndB} \to \R$ when it is chosen appropriately. In particular, we assume a posterior of the form
    \begin{multline}
        \density_{\networkParamsB}(\Vector|\CorVector) := \density_{(\networkParamsB_1,\networkParamsB_2)}(\Vector|\CorVector) := 
        \frac{e^{-\frac{1}{2} (\diffeo_{\networkParamsB_1}(\Vector) - \mathbf{m}_{\networkParamsB_2} (\CorVector))^\top \spdMatrix_{\networkParamsB_2}(\CorVector)^{-1} (\diffeo_{\networkParamsB_1}(\Vector)- \mathbf{m}_{\networkParamsB_2} (\CorVector)) } |\det(D_{\Vector} \diffeo_{\networkParamsB_1})|}{\sqrt{(2\pi)^\dimInd \det(\spdMatrix_{\networkParamsB_2})(\CorVector)}}
    \label{eq:paper-density-model-cond}
    \end{multline}
    where $\mathbf{m}_{\networkParamsB_2}: \R^\dimIndB\to\R^\dimInd$ is a learnable mean and $\spdMatrix_{\networkParamsB_2}:\R^\dimIndB\to\R^{\dimInd\times \dimInd}$ is a learnable diagonal matrix with positive entries, and use $\diffeo_{\networkParamsB_1}^{-1} \circ \mathbf{m}_{\networkParamsB_2}$ to generate posterior samples. 
    
    Then, using the data obtained through (i) or (ii) we build an RAE in the classical way \cite{diepeveen2024pulling}, by computing the Riemannian barycenter of the samples and performing tangent-space PCA \cite{diepeveen2025manifold} to obtain an orthonormal basis\footnote{Importantly, this PCA is carried out with respect to the $\ell^2$-inner product on the tangent space, as advocated in recent work \cite{diepeveen2025manifold}, rather than with respect to the metric tensor, which is the classical choice \cite{fletcher2004principal,diepeveen2025curvature}.}.

    \begin{remark}
        Although $\diffeo_{\networkParamsB_1}^{-1} \circ \mathbf{m}_{\networkParamsB_2}$ appears to solve the inverse problem, it typically overfits the training data. Consequently, when new data arrive, it is preferable to exploit the learned manifold structure for reconstruction instead of relying directly on $\diffeo_{\networkParamsB_1}^{-1} \circ\mathbf{m}_{\networkParamsB_2}$. We discuss ways in which this manifold structure can be directly exploited using the RAE in the next section.
    \end{remark}

\section{Beneficial Decoder Properties}\label{sec:decoder-properties} 

Once learned, the diffeomorphism $\diffeo_\networkParams$ and induced RAE can be exploited in downstream tasks with several beneficial properties. Recall the definition of the RAE decoder \cref{eq:rae-encoder-nf}: \begin{align*}
\RAEdecoder_\RAErelerror(\latentVector)& :=\exp_{\bar{\Vector}}^{\diffeo_\networkParams} \Bigl( \tangentbasis\latentVector\Bigr) = \diffeo_\networkParams^{-1}\left(\diffeo_\networkParams(\bar{\Vector}) + D_{\bar{\Vector}}\diffeo_\networkParams(\tangentbasis\latentVector)\right)\end{align*} where $\bar{\Vector}$ is a fixed base point and the columns of $\tangentbasis \in \R^{d \times \latentdim}$ with $d > \latentdim$ form an orthonormal basis for the tangent space $\mathcal{T}_{\bar{\Vector}}\R^d$. For an arbitrary diffeomorphism $\diffeo_\networkParams$, this decoder has a number of useful properties. In particular, note that it is naturally injective as it is the composition of an injective linear map and a bijective mapping. This allows for invertibility on its range, which is particularly attractive in applications such as inverse problems and Bayesian inference \cite{kothari2021trumpets}. Even more, this mapping enjoys several smoothness properties. We first recall the following definition. To state it, let $J_f(\Vector)$ denote the Jacobian of a map $f$ at a differentiable point $\Vector$.\footnote{In particular, note that $J_f(\cdot) = D_{(\cdot)}f$ as in \cref{sec:dd-rg}.}

\begin{definition}[Bi-Lipschitz and smooth decoder] \label{def:smoothness-decoder}
    We say that a function $\RAEdecoder : \R^r \rightarrow \R^d$ is $(m_1,m_2)$-bi-Lipschitz with $m_1,m_2> 0$ if  $$m_1\|\mathbf{p}-\mathbf{q}\|_2 \leq \|\RAEdecoder(\mathbf{p}) - \RAEdecoder(\mathbf{q})\|_2 \leq m_2\|\mathbf{p}-\mathbf{q}\|_2,\ \forall \mathbf{p},\mathbf{q} \in \R^{r}$$ Moreover, its Jacobian is $M$-Lipschitz if  $$\|J_{\RAEdecoder}(\mathbf{p}) - J_{\RAEdecoder}(\mathbf{q})\| \leq M\|\mathbf{p}-\mathbf{q}\|_2,\ \forall \mathbf{p},\mathbf{q} \in \R^{r}.$$
\end{definition}

Prior to showing the decoder's smoothness properties, we first discuss our choice of neural network parametrization.

\paragraph{Network Parametrization} The performance and expressiveness of the RAE is highly dependent on the choice of network architecture for the diffeomorphism $\diffeo_\networkParams$. In practice, we parametrize $\RAEdecoder_\RAErelerror$ via normalizing flows with additive coupling layers. More precisely, we consider \begin{align}
\diffeo_\networkParams(\Vector): = \phi^{(L)} \circ \dots \circ \phi^{(1)}(\Vector) \label{eq:rae-decoder-definition-tanh-layers}
\end{align} 
where each individual layer $\phi^{(i)}$ for $i \in [L]$ is given by 
$$
\phi^{(i)}(\stoVector) = f^{(i)} \circ \mathbf{V}^{(i)}(\stoVector)\ \text{with}\ f^{(i)}(\mathbf{z}) = [\mathbf{z}_1,\ \mathbf{z}_2 + g^{(i)}(\mathbf{z}_1)]^\top$$ where $\mathbf{z} =[\mathbf{z}_1,\ \mathbf{z}_2]^\top$ with $\mathbf{z}_1,\mathbf{z}_2 \in \R^{d/2}$, $\mathbf{V}^{(i)}$ is an invertible linear mapping, and $g^{(i)} : \R^{d/2} \rightarrow \R^{d/2}$ satisfies $$g^{(i)}(\mathbf{u}) = \left[\sum_{r=1}^n \alpha_{r,\ell}^{(i)} \tanh(\mathbf{u}_{\ell})^r\right]_{\ell=1}^{d/2}.$$ Here, $(\mathbf{V}^{(i)}, \alpha_{r,\ell}^{(i)})$ denote the learnable parameters of our diffeomorphism. This particular choice of diffeomorphism has two attributes that we highlight. The first is that the invertible linear layers allow for the map to be not necessarily volume-preserving, but have constant determinant $\Vector \mapsto |\mathrm{det}(D_{\Vector}\diffeo)|$ (as discussed in \cref{sec:framework}).  Moreover, the nonlinear bijective maps $f^{(i)}$ are volume-preserving, but have bounded derivatives that are explicitly controlled.

This particular parametrization shares many useful properties. In particular, the map $\latentVector \mapsto \RAEdecoder_\RAErelerror(\latentVector)$ is injective as $\diffeo_\networkParams$ is bijective, since each layer $\phi^{(i)}$ is bijective and $\tangentbasis$ has orthonormal columns. Moreover, we show that the RAE decoder given by the above network parametrization satisfies \cref{def:smoothness-decoder} with explicit constants. We provide a proof of this result in \cref{appx:decoder-smoothness}, along with an extension to more general diffeomorphisms.

\begin{proposition} \label{prop:decoder-smoothness} 
    There exists positive absolute constants $C_1,C_2$ such that the following holds. For each layer $i$, set $\underline{\sigma}_i := \sigma_{\min}((\mathbf{V}^{(L+1-i)})^{-1})$, $\overline{\sigma}_i := \sigma_{\max}((\mathbf{V}^{(L+1-i)})^{-1})$, $\tilde{\sigma}_{\min} :=\sigma_{\min}(D_{\bar{\Vector}}\diffeo_{\networkParams}\mathbf{U}_{\varepsilon})$, $\tilde{\sigma}_{\max} :=\|D_{\bar{\Vector}}\diffeo_{\networkParams}\mathbf{U}_{\varepsilon}\|$ and $B_i := C_1\max_{\ell \in [d/2]}\sum_{r=1}^\dimInd |\alpha^{(L+1-i)}_{r,\ell}|$. Then $\RAEdecoder_\RAErelerror$ with diffeomorphism $\diffeo_\networkParams$ as defined in \cref{eq:rae-decoder-definition-tanh-layers} is $(m_1,m_2)$-bi-Lipschitz and its Jacobian is $M$-Lipschitz where $$m_1 \geq\tilde{\sigma}_{\min} \prod_{i=1}^L\frac{\underline{\sigma}_i}{1+B_i},\ \quad\ m_2 \leq\tilde{\sigma}_{\max}\prod_{i=1}^L\overline{\sigma}_i(1+B_i),$$ and \begin{align*} 
    M & \leq C_2  \tilde{\sigma}_{\max}\sum_{\ell=1}^L\biggr[\prod_{i=\ell+1}^L\overline{\sigma}_i(1+B_i)\overline{\sigma}_{\ell} \biggr(\max_{j}\sum_{r=1}^{n}|\alpha^{(L+1-\ell)}_{r,j}|\biggr)\biggr(\prod_{k=1}^{\ell-1}\overline{\sigma}_k(1+B_k)\biggr)^2\biggr].
    \end{align*}
\end{proposition}

\subsection{Application to Inverse Problems}

Once learned, the RAE decoder can be used for downstream tasks that would benefit from a parametrization of the underlying data manifold. One such application would be in inverse problems, where the goal would be to reconstruct a signal $\stoVector_*$ from linear measurements $\CorVector = \forward\stoVector_* + \mathbf{n}$. As the RAE decoder $\RAEdecoder_\RAErelerror$ directly parametrizes an underlying data manifold, one could formulate reconstructing the signal via the following nonlinear least squares problem \begin{align}
    \min_{\mathbf{p} \in \R^{d_{\RAErelerror}}} \mathcal{L}(\mathbf{p}) := \frac{1}{2}\left\|\forward\RAEdecoder_\RAErelerror(\mathbf{p}) - \CorVector\right\|^2_2. \label{eq:inverseproblems-loss}
\end{align} This problem naturally incorporates the constraint that the reconstructed signal $\hat{\stoVector}$ should live on the data manifold and has been extensively used in the context of exploiting generative priors for inverse problems \cite{bora2018ambientgan}. It is also worthy to note that once $\RAEdecoder_\RAErelerror$ is learned via Riemannian AmbientFlow, it can be used for inverse problems in a \textit{forward-model agnostic} way, in that the forward operator $\forward$ need not have been the same forward operator that corrupted the data it was trained on.

We show that using our RAE decoder as a prior can yield recovery guarantees for inverse problems when minimizing \cref{eq:inverseproblems-loss} via gradient descent. To state our result, we require condition on the measurement matrix $\forward$ to ensure that the inverse problem is well-posed over the data manifold. In particular, we will require a slightly stronger version of the RIP that has been used in prior works on nonlinear recovery with generative priors \cite{hand2019global}.
\begin{definition}[Range-restricted isometry condition] \label{asmp:rric}
We say that $\forward \in \R^{m \times d}$ satisfies the range-restricted isometry condition (RRIC) with respect to $D : \R^r \rightarrow \R^d$ if the following holds for some $\delta \in (0,1)$: \begin{align*}
        & |( (\forward^\top\forward - \mathbf{I} )(\stoVector_1 - \stoVector_2),\stoVector_3 - \stoVector_4 )_2 | \leq \delta\|\stoVector_1 - \stoVector_2\|_2\|\stoVector_3-\stoVector_4\|_2,\ \forall \stoVector_1, \stoVector_2,\stoVector_3,\stoVector_4 \in \mathrm{range}(\RAEdecoder). 
    \end{align*} 
\end{definition} This condition ensures that the measurement operator acts like an isometry with respect to signals in the range of the RAE decoder $\RAEdecoder_\RAErelerror$, but is slightly stronger than the RIP as it requires control of the bilinear map $(\Vector,\Vector') \mapsto (\Vector, \forward \Vector')_2$ over the secant set $\mathrm{range}(\RAEdecoder_\RAErelerror)-\mathrm{range}(\RAEdecoder_\RAErelerror)$.

Equipped with these conditions, we show that when the decoder is sufficiently smooth and the measurement operator satisfies the RRIC, gradient descent with a sufficiently small step size will converge linearly up to a neighborhood of the true latent parameter whose radius depends on the amount of noise in the measurements.

\begin{theorem}[Recovery guarantees with RAE decoders] \label{thm:gd-convergence-ips}
    Suppose that $\CorVector=\forward\stoVector_* + \mathbf{n}$ where $\stoVector_* = \RAEdecoder_\RAErelerror(\mathbf{p}_*) \in \mathrm{range}(\RAEdecoder_\RAErelerror)$ for some $\mathbf{p}_* \in \R^{\latentdim}$ and $\RAEdecoder_\RAErelerror$ defined in \cref{eq:rae-encoder-nf}. Suppose $\forward$ satisfies the RRIC with respect to $\RAEdecoder_\RAErelerror$. Consider gradient descent with constant step size $\alpha > 0$ on the loss \cref{eq:inverseproblems-loss}: $\mathbf{p}_{t+1}:=\mathbf{p}_t - \alpha \nabla L(\mathbf{p}_t)$ for $t \geq 1$ and any initial iterate $\latentVector_1$. Then if \begin{align*}
        m_2M < &\ 2m_1^2,\  \delta  < \frac{m_1^2 - m_2M/2}{m_2^2}\ \text{and}\ 
        \alpha < \frac{m_1^2 - m_2M/2 - \delta m_2^2}{2m_2^4\|\forward^\top\forward\|^2}
    \end{align*} then there exists constants $\rho \in (0,1)$ and $\beta \in (0,\infty)$ that depend on $m_1,m_2,M,\delta,\forward,$ and $\alpha$ such that for all $t \geq 1$, we have $$\|\mathbf{p}_{t+1} - \mathbf{p}_*\|^2_2 \leq \rho ^{t} \|\mathbf{p}_1- \mathbf{p}_*\|^2_2 + \frac{\beta \|\forward^\top\mathbf{n}\|_2^2}{1-\rho}.$$
\end{theorem}

The proof of this result follows similar ideas to those in  \cite{latorre2019fast, nguyen2022provable}. In particular, convergence will follow from the following key lemma, which shows that under our conditions, the objective is smooth and satisfies a one-point strong convexity bound with an additional error term that depends on the noise:

\begin{lemma} \label{lem:strong-convexity}
    Under the conditions of \cref{thm:gd-convergence-ips}, we have that \begin{align}
        \|\nabla \mathcal{L}(\latentVector)\|^2_2 \leq 2m_2^4\|\forward^\top\forward\|^2\|\latentVector - \latentVector_*\|_2^2 + 2m_2^2\|\forward^\top\mathbf{n}\|_2^2 \label{eq:gradient-bound}
    \end{align} and 
\begin{equation}
\label{eq:one-point-strong-convexity} 
\begin{split}
  ( \latentVector - \latentVector_*, \nabla \mathcal{L}(\latentVector))_2 & \geq \frac{1}{2}\left(m_1^2 - \frac{m_2M}{2} - \delta m_2^2\right)\|\latentVector_t - \latentVector_*\|^2_2 - \frac{m_2^2\|\forward^\top\mathbf{n}\|_2^2}{2\left(m_1^2 - \frac{m_2M}{2} - \delta m_2^2\right)}.
\end{split}
\end{equation}
\end{lemma}
\begin{proof}[Proof of \cref{lem:strong-convexity}]
    Equation \cref{eq:gradient-bound} follows by noting \begin{align*}
        \nabla \mathcal{L}(\latentVector) & = J_{D_{\RAErelerror}}(\latentVector)^\top\forward^\top(\forward (D_{\RAErelerror}(\latentVector)-D_{\RAErelerror}(\latentVector_*)) + J_{D_{\RAErelerror}}(\latentVector)^\top\forward^\top\mathbf{n}
    \end{align*} and directly applying $\|\RAEdecoder_\RAErelerror(\latentVector) - \RAEdecoder_\RAErelerror(\latentVector_*)\|_2 \leq m_2\|\latentVector - \latentVector_*\|_2$,  $\|J_{\RAEdecoder_\RAErelerror}(\latentVector)\| \leq m_2$ for all $\latentVector \in \R^{\latentdim}$, along with the elementary bound $\|\mathbf{u} + \mathbf{v}\|_2^2 \leq 2\|\mathbf{u}\|^2_2 + 2\|\mathbf{v}\|_2^2$. For equation \cref{eq:one-point-strong-convexity}, let $\tilde{\nabla} \mathcal{L}(\latentVector) = J_{D_{\RAErelerror}}(\latentVector)^\top\forward^\top(\forward (D_{\RAErelerror}(\latentVector)-D_{\RAErelerror}(\latentVector_*))$ so that $\nabla \mathcal{L}(\latentVector) = \tilde{\nabla}\mathcal{L}(\latentVector) + J_{D_{\RAErelerror}}(\latentVector)^\top\forward^\top\mathbf{n}$. Then observe that \begin{align}
       ( \latentVector-\latentVector_*, \tilde{\nabla} \mathcal{L}(\latentVector))_2 \nonumber & = ( \latentVector-\latentVector_*, J_{\RAEdecoder_\RAErelerror}(\latentVector)^\top\forward^\top\forward(\RAEdecoder_\RAErelerror(\latentVector)-\RAEdecoder_\RAErelerror(\latentVector_*)))_2 \nonumber\\
        & = ( J_{\RAEdecoder_\RAErelerror}(\latentVector)(\latentVector-\latentVector_*), \RAEdecoder_\RAErelerror(\latentVector)-\RAEdecoder_\RAErelerror(\latentVector_*))_2 \nonumber\\
        & + ( J_{\RAEdecoder_\RAErelerror}(\latentVector)(\latentVector-\latentVector_*), (\forward^\top\forward-\mathbf{I} )(\RAEdecoder_\RAErelerror(\latentVector)-\RAEdecoder_\RAErelerror(\latentVector_*)))_2. \label{eq:one-point-starting}
    \end{align} For the first term, note that for any $\latentVector,\mathbf{q} \in \R^{\latentdim}$, we have that by the Fundamental Theorem of Calculus, \begin{align}
        & ( \RAEdecoder_\RAErelerror(\latentVector) - \RAEdecoder_\RAErelerror(\mathbf{q}), J_{\RAEdecoder_\RAErelerror}(\latentVector)(\latentVector-\mathbf{q}))_2 \nonumber \\
        & = \|\RAEdecoder_\RAErelerror(\latentVector)-\RAEdecoder_\RAErelerror(\mathbf{q})\|^2_2  -  ( \RAEdecoder_\RAErelerror(\latentVector) - \RAEdecoder_\RAErelerror(\mathbf{q}), \RAEdecoder_\RAErelerror(\latentVector) - \RAEdecoder_\RAErelerror(\mathbf{q})-J_{\RAEdecoder_\RAErelerror}(\latentVector)(\latentVector-\mathbf{q}))_2 \nonumber\\
        & =\|\RAEdecoder_\RAErelerror(\latentVector)-\RAEdecoder_\RAErelerror(\mathbf{q})\|^2_2 \nonumber  -  \int_0^1( \RAEdecoder_\RAErelerror(\latentVector) - \RAEdecoder_\RAErelerror(\mathbf{q}),(J_{\RAEdecoder_\RAErelerror}(\latentVector + t(\latentVector-\mathbf{q})) \nonumber -J_{\RAEdecoder_\RAErelerror}(\latentVector))(\latentVector-\mathbf{q}))_2\mathrm{d}t \nonumber\\
        & \geq m_1^2\|\latentVector-\mathbf{q}\|^2 - m_2M\|\latentVector-\mathbf{q}\|_2^2\int_0^1(1-t)\mathrm{d}t \label{eq:smoothness-bd}\\
        & = \left(m_1^2 - \frac{m_2M}{2}\right)\|\latentVector-\mathbf{q}\|^2_2 \label{eq:first-term-one-point}
    \end{align} where we used the fact that $D_{\varepsilon}$ is $(m_1,m_2)$-bi-Lipschitz and $J_{D_{\varepsilon}}$ is $M$-Lipschitz in \cref{eq:smoothness-bd}. For the second term, note that for any small $t > 0$, we have by the RRIC, \begin{align*}
        |( D_{\varepsilon}(&\latentVector+t(\latentVector -\latentVector_*))  - \RAEdecoder_\RAErelerror(\latentVector),(\forward^\top\forward-\mathbf{I})(\RAEdecoder_\RAErelerror(\latentVector)-\RAEdecoder_\RAErelerror(\latentVector_*)))_2| \\
        & \leq \delta\|D_{\varepsilon}(\latentVector+t(\latentVector-\latentVector_*))-\RAEdecoder_\RAErelerror(\latentVector)\|_2\|\RAEdecoder_\RAErelerror(\latentVector)-\RAEdecoder_\RAErelerror(\latentVector_*)\|_2\\
        & \leq \delta m_2^2 t\|\latentVector-\latentVector_*\|^2_2
    \end{align*} where we used the fact that $D_{\varepsilon}$ is $m_2$-Lipschitz in the last inequality. Since $J_{\RAEdecoder_\RAErelerror}(\latentVector)(\latentVector-\latentVector_*) = \frac{\mathrm{d}}{\mathrm{d}t} D_{\varepsilon}(\latentVector+t(\latentVector-\latentVector_*))\rvert_{t=0} = \lim_{t\rightarrow0}t^{-1}(D_{\varepsilon}(\latentVector+t(\latentVector-\latentVector_*)) - \RAEdecoder_\RAErelerror(\latentVector))$, we have that dividing by $t > 0$ and taking the limit $t \rightarrow 0$ yields \begin{align}
        ( J_{\RAEdecoder_\RAErelerror}(\latentVector)(\latentVector-\latentVector_*), (\forward^\top\forward &-\mathbf{I})(\RAEdecoder_\RAErelerror(\latentVector)-\RAEdecoder_\RAErelerror(\latentVector_*)))_2  \geq -\delta m_2^2\|\latentVector-\latentVector_*\|^2_2. \label{eq:second-term-one-point}
    \end{align}
 Using equations \cref{eq:first-term-one-point} and \cref{eq:second-term-one-point} in \cref{eq:one-point-starting} yields $$( \latentVector-\latentVector_*, \tilde{\nabla} \mathcal{L}(\latentVector))_2 \geq \left(m_1^2 - \frac{m_2M}{2} - \delta m_2^2\right)\|\latentVector-\latentVector_*\|^2_2.$$ For the remaining error term, note that \begin{align*}
     |( \latentVector - \latentVector_*,J_{D_{\RAErelerror}}(\latentVector)^\top\forward^\top\mathbf{n})_2| \leq m_2\|\latentVector-\latentVector_*\|_2\|\forward^\top\mathbf{n}\|_2.
 \end{align*} Setting $m_{\delta}:=m_1^2 - \frac{m_2M}{2} - \delta m_2^2$, we obtain \begin{align}
     ( \latentVector - \latentVector_*, \nabla \mathcal{L}(\latentVector))_2 \geq m_{\delta}\|\latentVector-\latentVector_*\|_2^2 - m_2\|\forward^\top\mathbf{n}\|_2 \|\latentVector-\latentVector_*\|_2. \label{eq:last-one-point-bound-with-noise}
 \end{align} Recall Young's inequality: for any $\mu > 0$, we have $ab \leq \mu a^2/2 + b^2/(2\mu)$. Substituting $\mu = m_{\delta}$, $a = \|\latentVector-\latentVector_*\|_2$ and $b = m_2\|\forward^\top\mathbf{n}\|_2$ yields $m_2\|\forward^\top\mathbf{n}\|_2\|\latentVector-\latentVector_*\|_2 \leq m_{\delta}\|\latentVector-\latentVector_*\|_2^2/2 + m_2^2\|\forward^\top\mathbf{n}\|_2^2/(2m_{\delta})$. Applying this bound to \cref{eq:last-one-point-bound-with-noise}, we obtain \cref{eq:one-point-strong-convexity}.
\end{proof}

We now proceed with the proof of \cref{thm:gd-convergence-ips}.

\begin{proof}[Proof of \cref{thm:gd-convergence-ips}]
For any $t \geq 1,$ we have that the iterates of gradient descent satisfy \begin{align*}
   &  \|\latentVector_{t+1}-\latentVector_*\|_2^2  =  \|\latentVector_t - \alpha \nabla \mathcal{L}(\latentVector_t) - \latentVector_*\|^2_2 \\
    & = \|\latentVector_t-\latentVector_*\|^2_2 + \alpha^2 \|\nabla \mathcal{L}(\latentVector_t)\|^2_2 - 2\alpha ( \latentVector_t - \latentVector_*, \nabla \mathcal{L}(\latentVector_t))_2.
\end{align*} If we apply both bounds from \cref{lem:strong-convexity}, setting $m_{\delta}:=m_1^2 - \frac{m_2M}{2} - \delta m_2^2$ we see that \begin{align*}
     \|\latentVector_{t+1}-\latentVector_*\|_2^2 
    & \leq \|\latentVector_t - \latentVector_*\|_2^2 +  2\alpha^2 m_2^4\|\forward^\top\forward\|^2\|\latentVector - \latentVector_*\|_2^2 + 2\alpha^2m_2^2\|\forward^\top\mathbf{n}\|_2^2 \\
    & \quad \quad - \alpha m_{\delta}\|\latentVector_t - \latentVector_*\|_2^2 + \frac{\alpha m_2^2\|\forward^\top\mathbf{n}\|_2^2}{m_{\delta}} \\
    & = \biggr(1 -\alpha m_{\delta} + 2\alpha^2 m_2^4\|\forward^\top\forward\|^2 \biggr)\|\latentVector_t - \latentVector_*\|^2_2 + \biggr(2\alpha^2m_2^2 + \frac{\alpha m_2^2}{m_{\delta}}\biggr)\|\forward^\top\mathbf{n}\|_2^2\\
    & =: \rho \|\latentVector_t - \latentVector_*\|^2_2 + \beta \|\forward^\top\mathbf{n}\|_2^2.
\end{align*} Choosing $\delta$ such that $\delta m_2^2 < m_1^2 - \frac{m_2M}{2}$ so that $m_{\delta} > 0$ and $\alpha$ such that $$\alpha < \frac{m_{\delta}}{2m_2^4\|\forward^\top\forward\|^2}$$ guarantees that $\rho \in (0,1)$. Unrolling this recursion and summing a geometric series yields \begin{align*}
    \|\latentVector_{t+1}-\latentVector_*\|_2^2 & \leq  \rho^{t} \|\latentVector_1 - \latentVector_*\|_2^2 + \beta \|\forward^\top\mathbf{n}\|_2^2 \cdot \sum_{i=0}^{t-1}\rho^i \\
    & = \rho^{t} \|\latentVector_1 - \latentVector_*\|_2^2 + \beta \|\forward^\top\mathbf{n}\|_2^2 \cdot \frac{1-\rho^{t}}{1-\rho} 
    \leq \rho^{t} \|\latentVector_1 - \latentVector_*\|_2^2 + \frac{\beta \|\forward^\top\mathbf{n}\|_2^2}{1-\rho}
\end{align*} as desired.
\end{proof}

\section{Experiments}


We conduct two sets of experiments to study the behavior and practical performance of Riemannian AmbientFlow. First, we evaluate the method on controlled synthetic data to develop intuition and identify best practices for training and model selection. Using these insights, we then turn to real-world data with the MNIST dataset, where we examine in more detail which aspects of our approach work effectively and where its current limitations become apparent.


\subsection{A Corrupted Sinusoid}


We first consider a synthetic dataset consisting of 1000 corrupted samples and 50 clean reference samples, corresponding to a 5\% clean-to-corrupt data ratio. The underlying data points are generated from a sinusoidal curve embedded in $\R^5$ and subsequently mapped to $\R^3$ via a random Gaussian matrix\footnote{With high probability, this random projection constitutes an embedding, allowing us to visualize the data in three dimensions.}, after which Gaussian noise with standard deviation $\sigma = 0.1$ is added. Importantly, our modeling assumption -- that the distribution resembles a deformed Gaussian -- closely matches the true generative process. Consequently, this setting primarily serves as a sanity check for our method.

For the loss function, $M=10$ samples are used in $\mathcal{L}_{\text{VLB}}$ and regularization parameters are set to $\lambda = 0.1$ and $\mu = 1.0$ for the low-rank penalty and the reference-sample term, respectively.

\begin{figure}[h!]
  \centering
  \begin{subfigure}{0.24\textwidth}
    \centering
    \includegraphics[width=\textwidth]{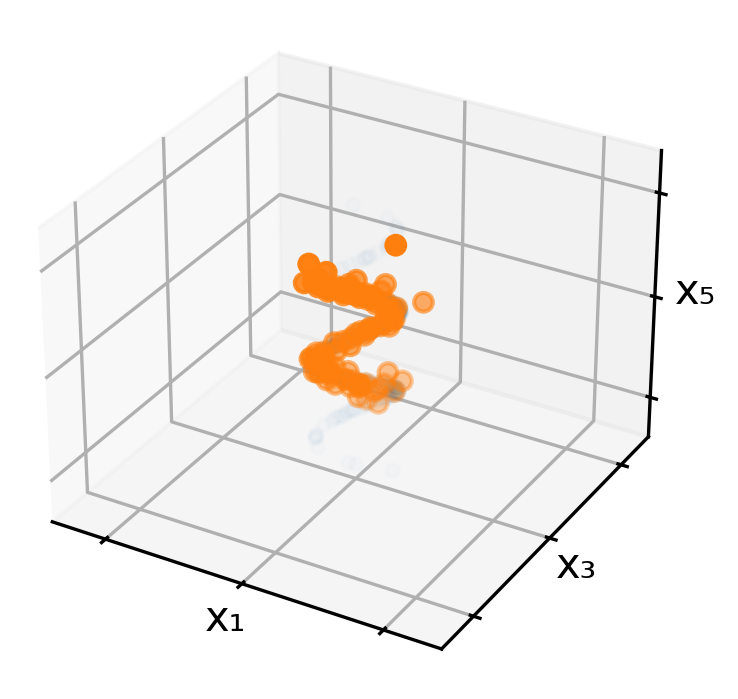}
    \caption{Clean samples}
    \label{fig:x135}
  \end{subfigure}
  \hfill
  \begin{subfigure}{0.24\textwidth}
    \centering
    \includegraphics[width=\textwidth]{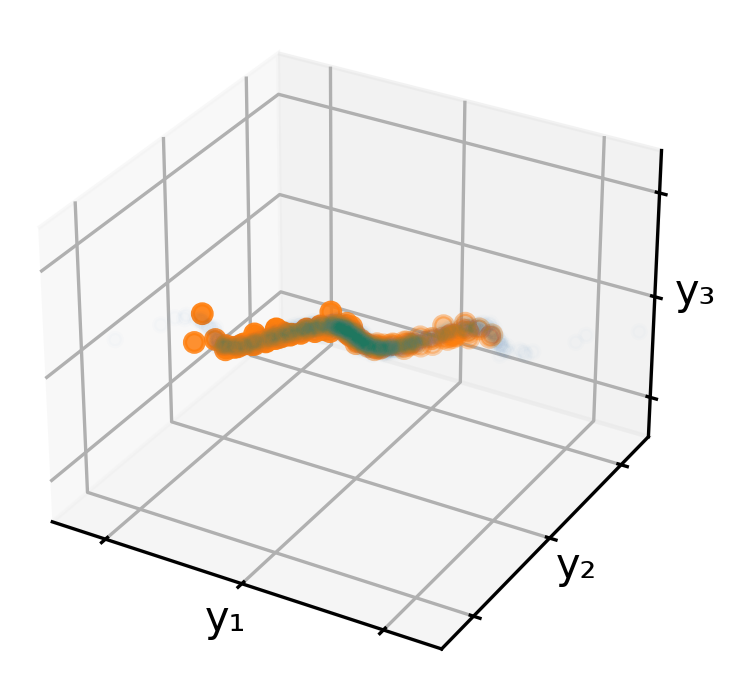}
    \caption{Corrupted samples}
    \label{fig:x345}
  \end{subfigure}
  \hfill
  \begin{subfigure}{0.24\textwidth}
    \centering
    \includegraphics[width=\textwidth]{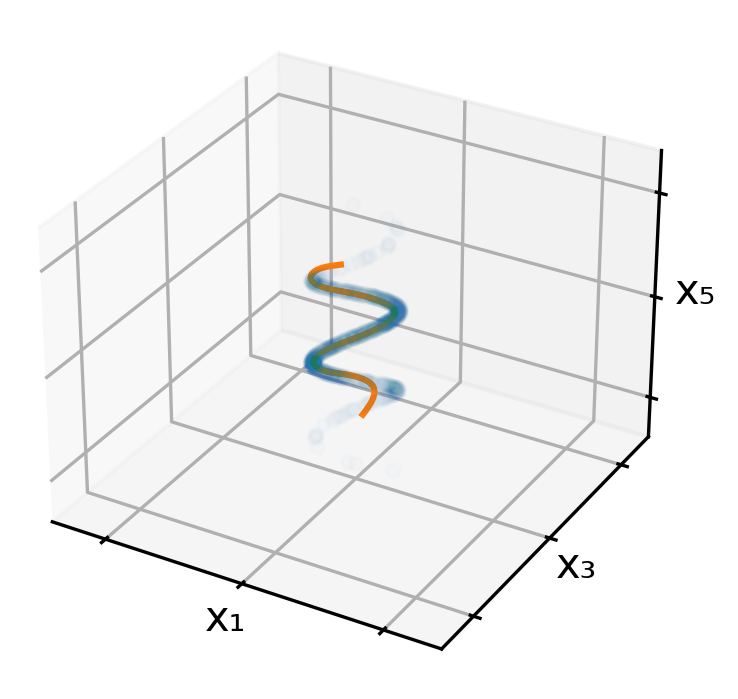}
    \caption{Clean manifold}
    \label{fig:y_clean}
  \end{subfigure}
  \hfill
  \begin{subfigure}{0.24\textwidth}
    \centering
    \includegraphics[width=\textwidth]{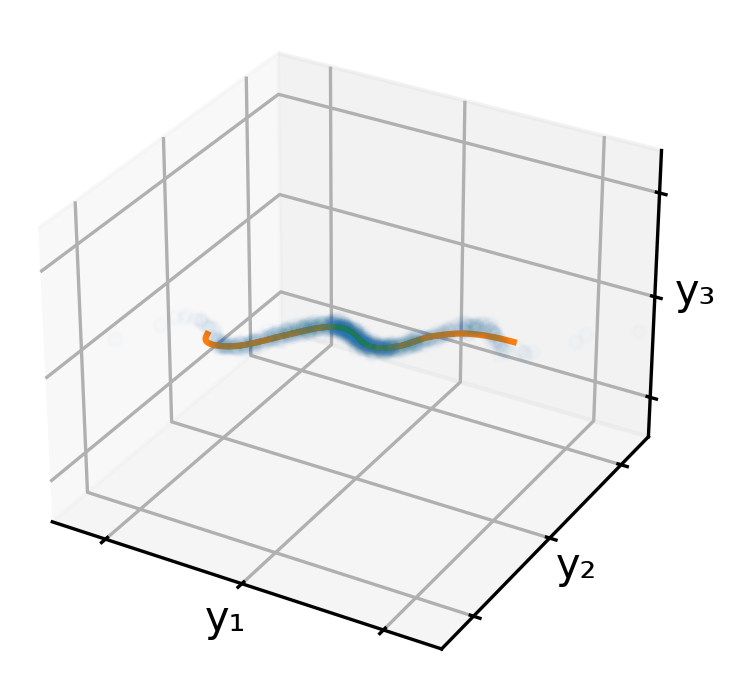}
    \caption{Corrupted manifold}
    \label{fig:y_corr}
  \end{subfigure}
  \caption{After 500 iterations of minimizing the Riemannian AmbientFlow problem \cref{eq:full-rie-ambientflow-loss} we have learned both a generative model and a smooth manifold that matches both the clean and corrupted data distribitions.}
  \label{fig:sinusoid_views}
\end{figure}

For the model architectures, the learnable diffeomorphism $\diffeo_\networkParams$ underlying the prior $\density_\networkParams$ follows the parametrization introduced in \cref{eq:rae-decoder-definition-tanh-layers}, employing invertible linear layers represented as a product of learnable upper and lower triangular matrices. The posterior $\density_\networkParamsB$ is defined as in \cref{eq:paper-density-model-cond} through learnable mean $\mathbf{m}_{\networkParamsB}$ and learnable diagonal covariance $\spdMatrix_{\networkParamsB}$ and fixed identity diffeomorphism. The two learnable mappings are parametrized as $(\mathbf{m}_{\networkParamsB} (\CorVector), \operatorname{diag} (\spdMatrix_{\networkParamsB}(\CorVector))) := f_\networkParams (\CorVector)$ where $f_\networkParams:\R^3\to \R^{10}$ is a learnable neural network composed of a linear embedding from $\R^3$ into $\R^{10}$, followed by a single ResNet block with hidden layer width 8. 

\begin{figure}[h!]
\centering
\begin{subfigure}{0.24\textwidth}
\centering
\includegraphics[width=\textwidth]{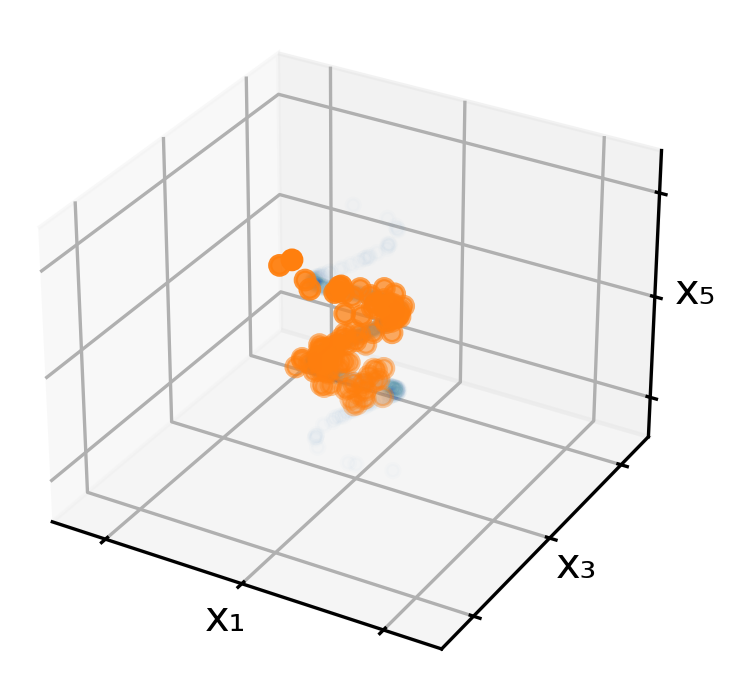}
\caption{Clean samples}
\label{fig:sinusoid_250_x135}
\end{subfigure}
\hfill
\begin{subfigure}{0.24\textwidth}
\centering
\includegraphics[width=\textwidth]{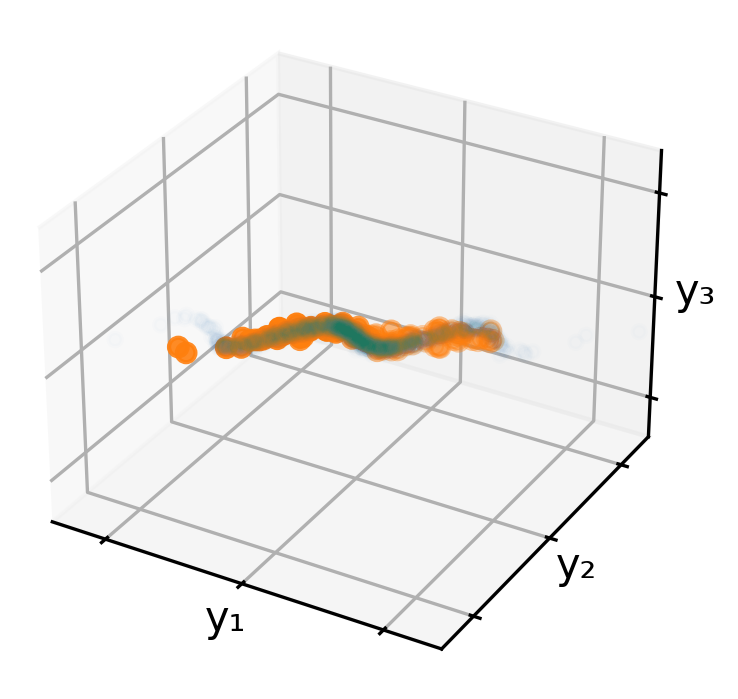}
\caption{Corrupted samples}
\label{fig:sinusoid_250_x345}
\end{subfigure}
\hfill
\begin{subfigure}{0.24\textwidth}
\centering
\includegraphics[width=\textwidth]{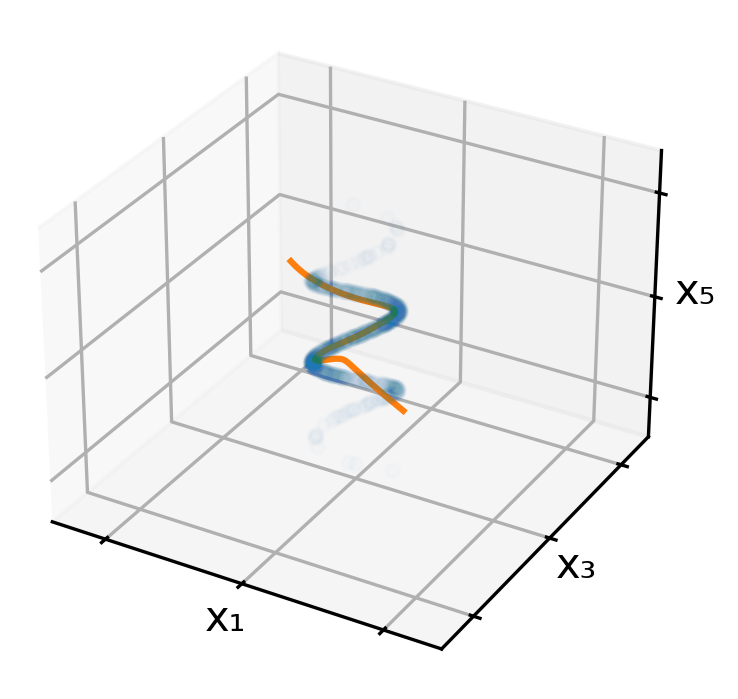}
\caption{Clean manifold}
\label{fig:sinusoid_250_y_clean}
\end{subfigure}
\hfill
\begin{subfigure}{0.24\textwidth}
\centering
\includegraphics[width=\textwidth]{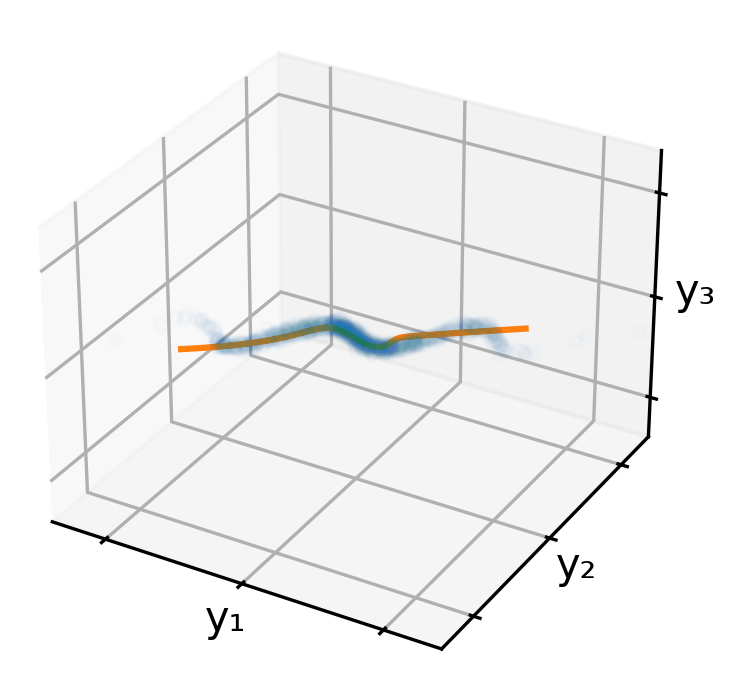}
\caption{Corrupted manifold}
\label{fig:sinusoid_250_y_corr}
\end{subfigure}
\caption{After 250 iterations of minimizing the Riemannian AmbientFlow problem \cref{eq:full-rie-ambientflow-loss} we have learned a generative model that matches the corrupted data distribition, but is still finetuning the model that approximates clean data distribution. This is also noticeable in the quality of the learned manifolds -- both for the clean data manifold as the corrupted data manifold --, which are not quite as accurate as the ones from Figure~\ref{fig:sinusoid_views}.}
\label{fig:sinusoid_views_250}
\end{figure}

Both components are trained using the Adam optimizer with a learning rate of $10^{-3}$ for 500 iterations and the results shown in \cref{fig:sinusoid_views} showcase that the model is indeed able to jointly learn a generative model and a manifold representation of the clean data distribution, where the reference data points are used to construct the RAE with latent dimension 1, whose decoder is used to generate the manifold. Having said that, several aspects are worth highlighting:

\paragraph{Sufficient Iterations}
By considering the results at 250 iterations in \cref{fig:sinusoid_views_250}, we can see that effectively the embedding network $f_\networkParamsB$ underlying the posterior density $\density_\networkParamsB$ is trained first, followed by the prior density $\density_\networkParams$. This training behavior requires notably more iterations to learn the prior than would be needed for direct training on the clean data. Nevertheless, after sufficient optimization, both a well-calibrated generative model and a coherent learned manifold are obtained.

\begin{figure}[h!]
\centering
\begin{subfigure}{0.24\textwidth}
\centering
\includegraphics[width=\textwidth]{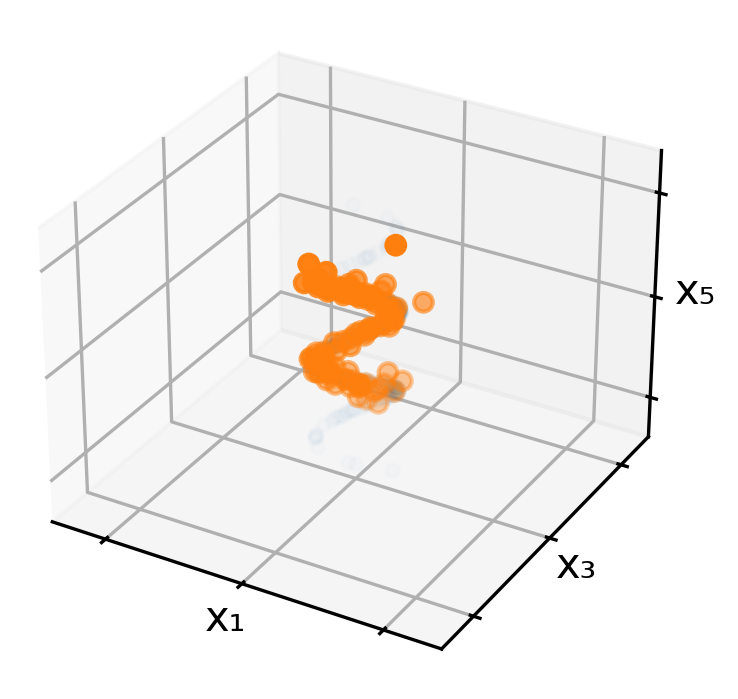}
\caption{Clean samples}
\label{fig:sinusoid_lam0_x135}
\end{subfigure}
\hfill
\begin{subfigure}{0.24\textwidth}
\centering
\includegraphics[width=\textwidth]{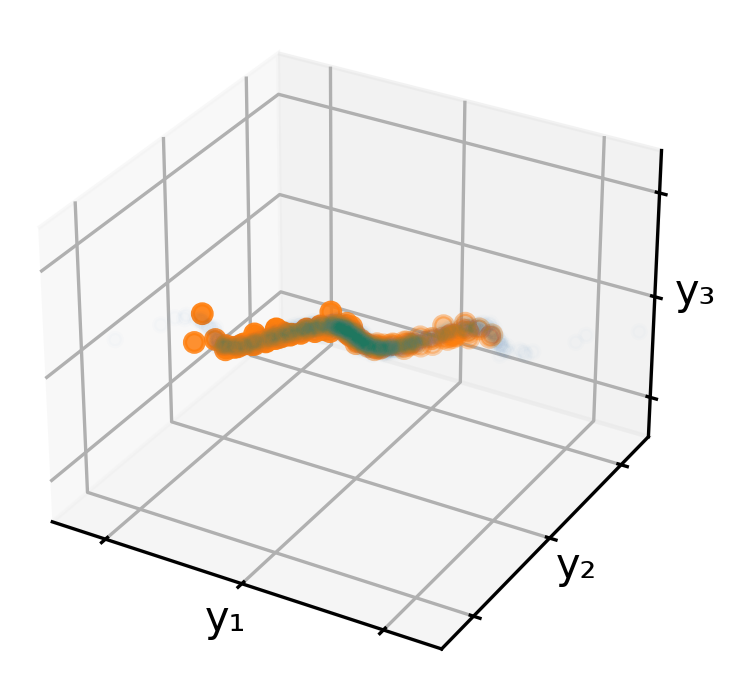}
\caption{Corrupted samples}
\label{fig:sinusoid_lam0_x345}
\end{subfigure}
\hfill
\begin{subfigure}{0.24\textwidth}
\centering
\includegraphics[width=\textwidth]{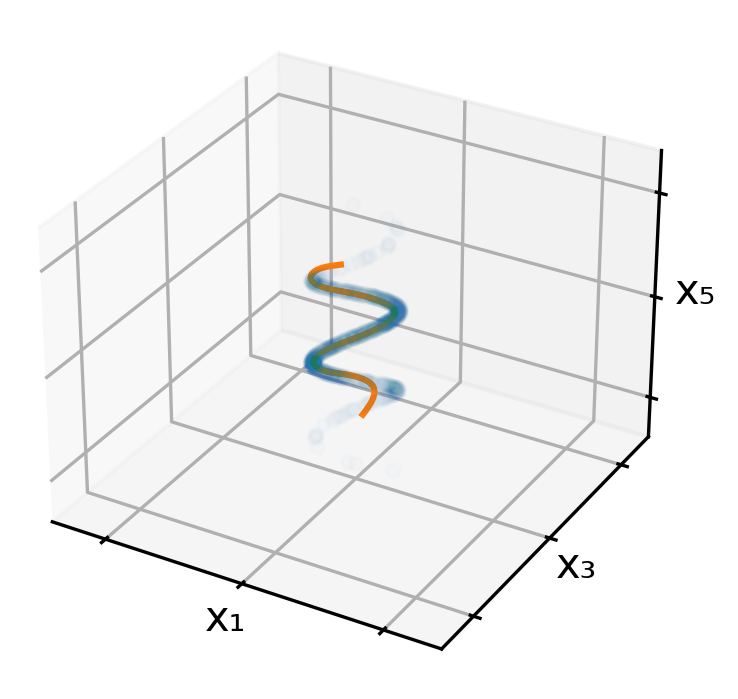}
\caption{Clean manifold}
\label{fig:sinusoid_lam0_y_clean}
\end{subfigure}
\hfill
\begin{subfigure}{0.24\textwidth}
\centering
\includegraphics[width=\textwidth]{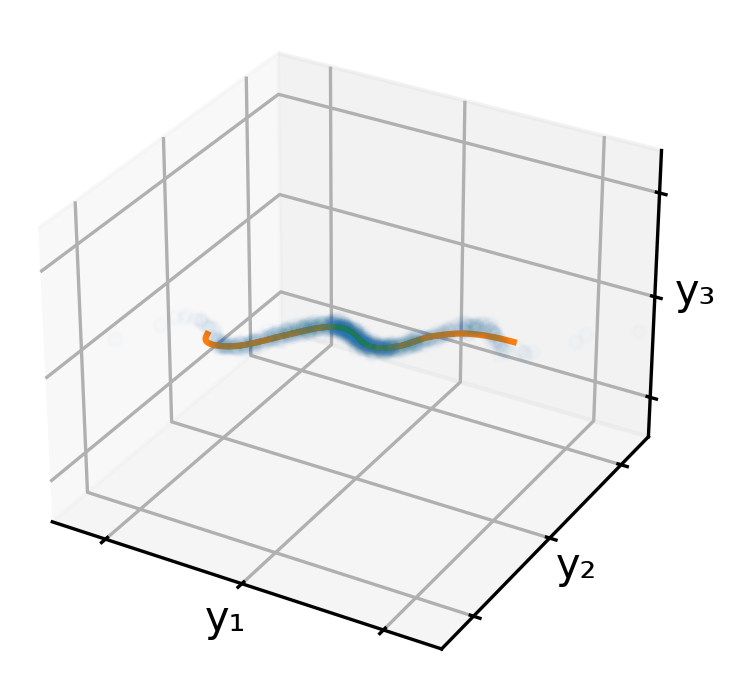}
\caption{Corrupted manifold}
\label{fig:sinusoid_lam0_y_corr}
\end{subfigure}
\caption{After 500 iterations of minimizing the Riemannian AmbientFlow problem \cref{eq:full-rie-ambientflow-loss} with $\lambda = 0$ we have learned both a generative model and a smooth manifold that matches both the clean and corrupted data distribitions.}
\label{fig:sinusoid_views_lam0}
\end{figure}

\begin{figure}[h!]
\centering
\begin{subfigure}{0.24\textwidth}
\centering
\includegraphics[width=\textwidth]{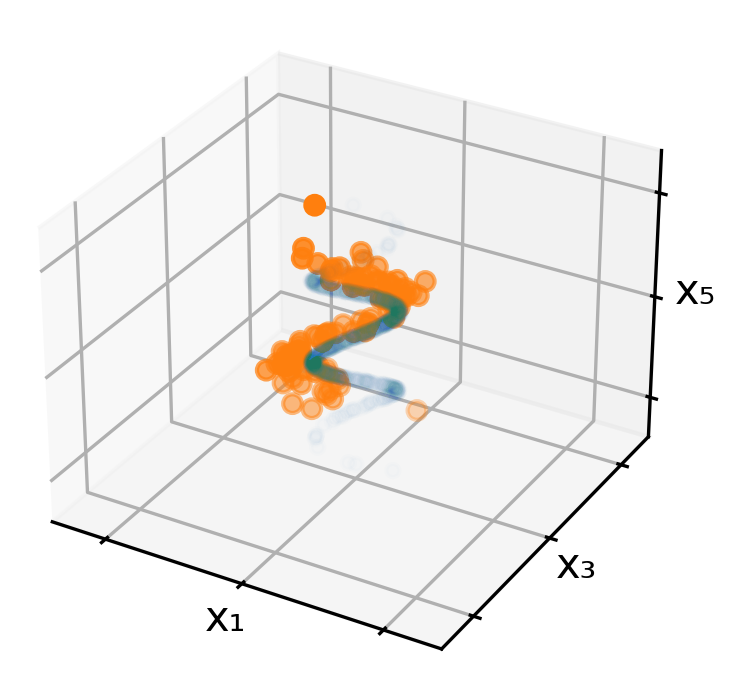}
\caption{Clean samples}
\label{fig:sinusoid_mu0_x135}
\end{subfigure}
\hfill
\begin{subfigure}{0.24\textwidth}
\centering
\includegraphics[width=\textwidth]{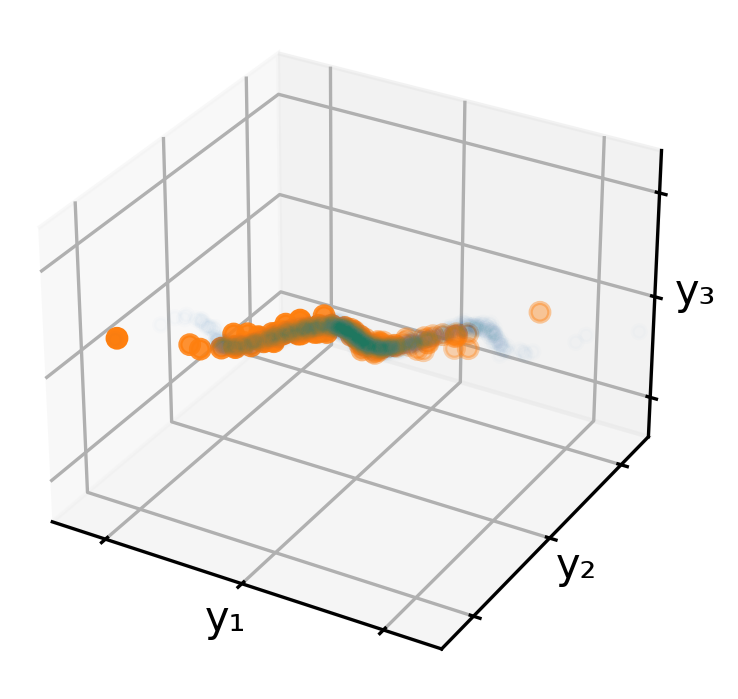}
\caption{Corrupted samples}
\label{fig:sinusoid_mu0_x345}
\end{subfigure}
\hfill
\begin{subfigure}{0.24\textwidth}
\centering
\includegraphics[width=\textwidth]{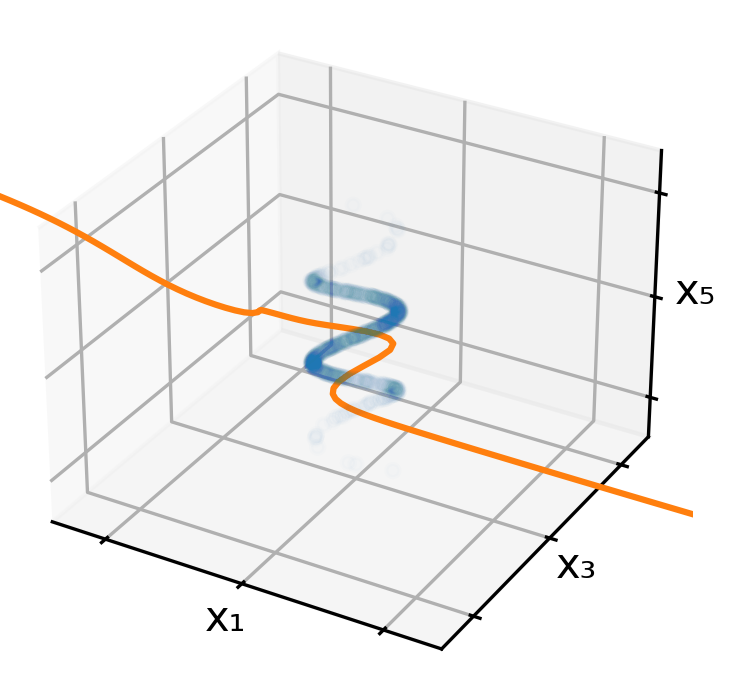}
\caption{Clean manifold}
\label{fig:sinusoid_mu0_y_clean}
\end{subfigure}
\hfill
\begin{subfigure}{0.24\textwidth}
\centering
\includegraphics[width=\textwidth]{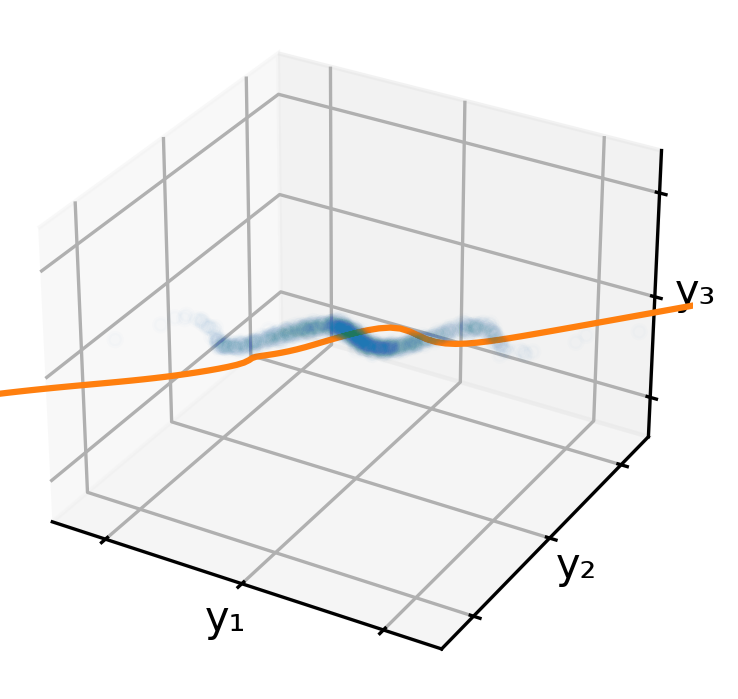}
\caption{Corrupted manifold}
\label{fig:sinusoid_mu0_y_corr}
\end{subfigure}
\caption{After 500 iterations of minimizing the Riemannian AmbientFlow problem \cref{eq:full-rie-ambientflow-loss} with $\mu = 0$ we have learned a generative model that matches the corrupted data distribition, but does not match the clean data distribution. Instead the distribution is offset in the null space of the forward operator. This is also noticeable in the quality of the learned manifolds -- both for the clean data manifold as the corrupted data manifold --, which are do not even come close to the ones learned in Figure~\ref{fig:sinusoid_views}.}
\label{fig:sinusoid_views_mu0}
\end{figure}

\paragraph{Sufficient Regularization}
Empirically, there is no observable difference between including or omitting the low-rank regularizer (\cref{fig:sinusoid_views_lam0}). In contrast, removing the reference samples causes the learned distribution to shift along the null space of the corruption operator (\cref{fig:sinusoid_views_mu0}). In this setting, the model captures the corrupted data distribution well but fails to recover a meaningful distribution on the clean data, resulting in degraded reconstructions of both the clean and corrupted manifolds.

\subsection{Blurred and Noisy MNIST}
\begin{figure}[h] 
\centering
\includegraphics[width=0.75\textwidth]{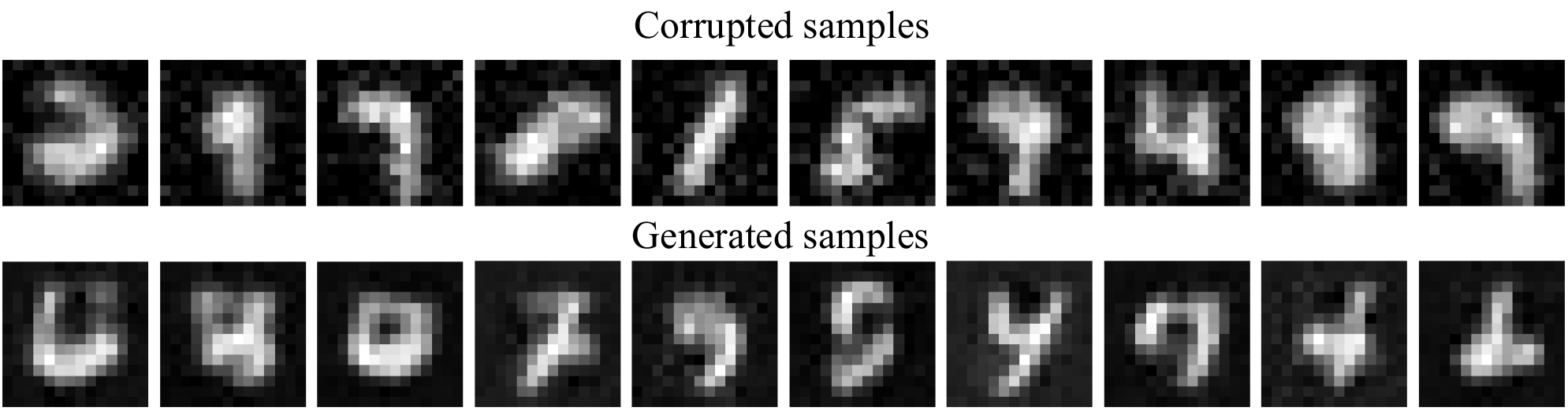}
\caption{Top row: $10$ randomly selected corrupted training images. Bottom row: $10$ randomly generated samples from our learned prior $p_{\networkParams}$.}
\label{fig:generated-samples}
\end{figure}

We now consider a dataset of corrupted MNIST digits. In particular, we will consider downsampled $14 \times 14$ MNIST digits corrupted by a Gaussian blur operator along with additive Gaussian noise. We sample $55000$ images from the MNIST training set and corrupt them using a fixed blur operator $\mathbf{A}$ with kernel size of $9$ and a $\sigma_{\mathrm{ker}} =1.5$. The additive noise has standard deviation $\sigma = 0.05$. Similar to the previous experiment, $100$ of these images will be used as clean reference samples. We also omit the use of the low-rank regularizer based on the higher dimensionality of this problem and the lack of observable difference in performance with the use of this regularizer in the previous synthetic dataset. In terms of the diffeomorphism's parametrization, we employ the network discussed in \cref{sec:decoder-properties}, where the linear layers are invertible convolutional layers. We train the model by optimizing \cref{eq:full-rie-ambientflow-loss} with $(\lambda,\mu) = (0,100)$ for $480$ epochs\footnote{We observed empirically that, similar to the previous experiment, a sufficient number of epochs was necessary to achieve good generation quality. This also impacted the quality of the RAE used in the next subsection.}, a batch size of $250$, and a learning rate of $10^{-3}$. We show generated samples from our learned model in \cref{fig:generated-samples}. While Riemannian Ambientflow is trained on corrupted images, the generates samples exhibit less noise and are slightly sharper than the blurry corrupted samples.

\begin{figure}[h] 
\centering
\includegraphics[width=0.75\textwidth]{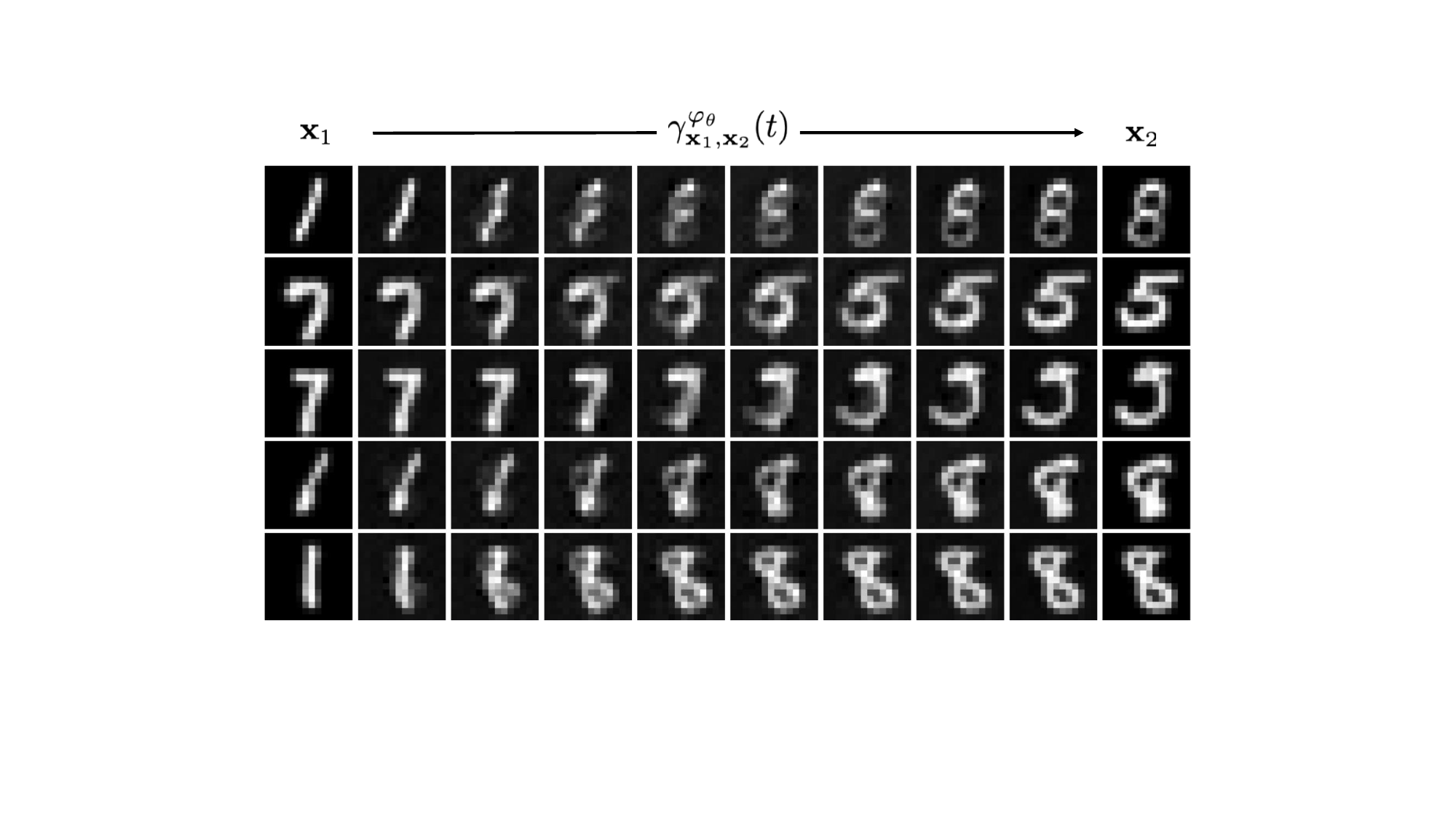}
\caption{Points along the geodesic $\gamma_{\mathbf{x}_1,\mathbf{x}_2}^{\varphi_{\networkParams}}(t)$ from our learned manifold for $t \in \{0, 1/9, 2/9, \dots, 8/9, 1\}$, interpolating from one sample $\mathbf{x}_1$ to another $\mathbf{x}_2$. Each row corresponds to a different pair $(\mathbf{x}_1,\mathbf{x}_2)$ and points along the geodesic $\gamma_{\mathbf{x}_1,\mathbf{x}_2}^{\varphi_{\networkParams}}(t)$.}
\label{fig:clean-geodesics}
\end{figure}

Given our learned prior, we can construct a pullback manifold to understand the learned geometry from our model. For example, we can analyze geodesics of the manifold by interpolating between two clean images $\mathbf{x}_1,\mathbf{x}_2$ from the dataset via \cref{eq:thm-geodesic-remetrized}, i.e.,
\begin{align*}
    \gamma_{\mathbf{x}_1,\mathbf{x}_2}^{\varphi_{\networkParams}}(t) = \varphi_{\networkParams}^{-1}((1-t)\varphi_{\networkParams}(\mathbf{x}_1) + t \varphi_{\networkParams}(\mathbf{x}_2)),\ \quad\ t \in [0,1].
\end{align*} We see in \cref{fig:clean-geodesics} that geodesics between two clean samples interpolates between different classes in semantically meaningful ways. These two results illustrate a benefit of our approach where we are able to simultaneously learn a generative model and a manifold from corrupted data.

\begin{figure}[h] 
\centering
\includegraphics[width=0.75\textwidth]{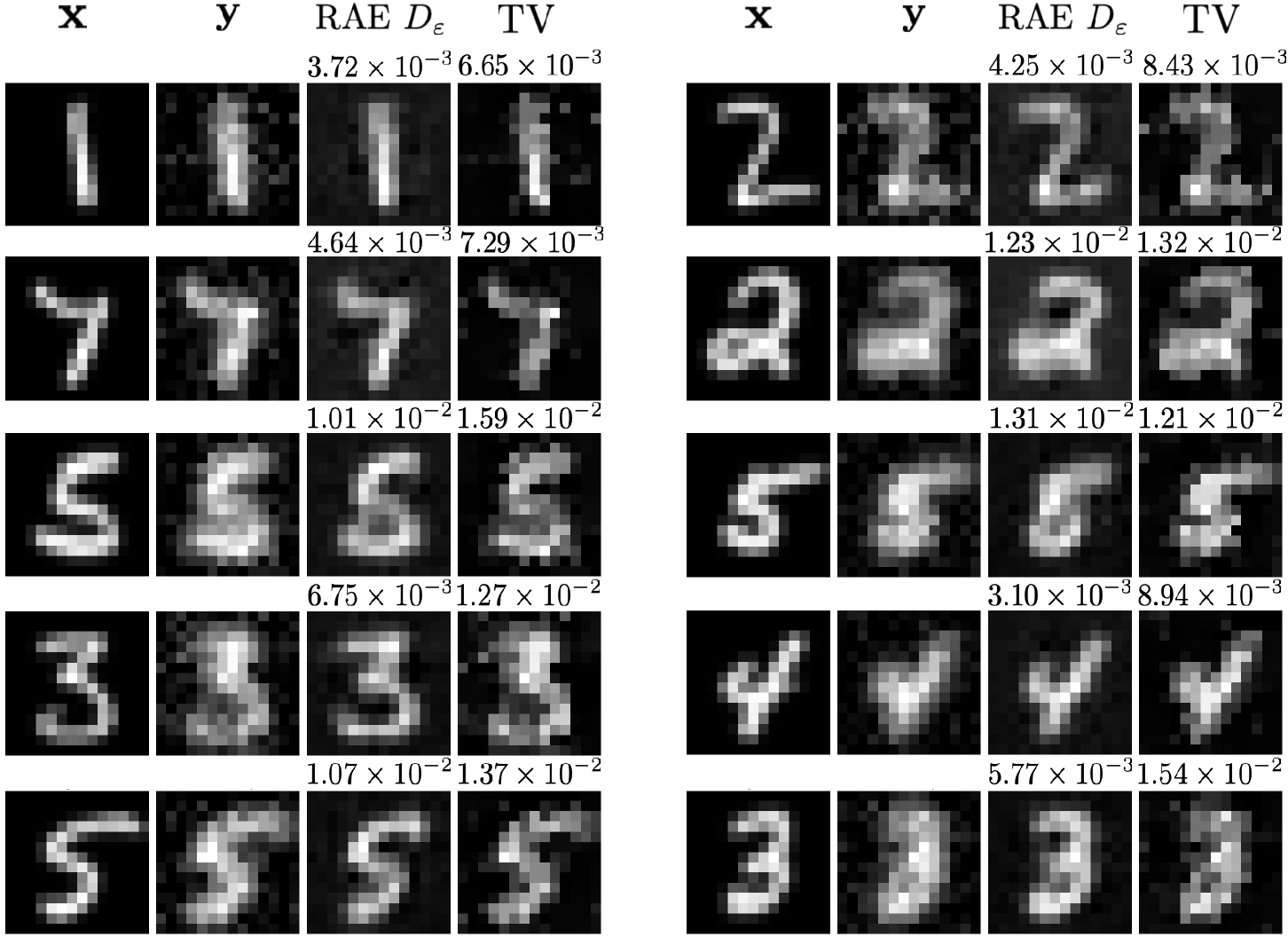}
\caption{We showcase reconstruction results on solving inverse problems using a learned RAE decoder as a prior. For each column, $\mathbf{x}$ denotes a test image, $\mathbf{y}$ denotes a corrupted measurement example, RAE $D_{\varepsilon}$ is the reconstruction with our method and TV is the TV reconstruction based on the iterations \cref{eq:tv-alg}. We also include the MSE above each reconstruction. The average MSE for the RAE over these $10$ test images was $7.43 \times 10^{-3}$ while for TV the average MSE was $1.14 \times 10^{-2}$.}
\label{fig:inv-probs}
\end{figure}

\paragraph{Exploiting the Decoder Prior} We build the RAE $D_{\varepsilon}$ from the learned diffeomorphism $\varphi_{\networkParams}$ to use as a prior for inverse problems. Specifically, the basis $\mathbf{U}_{\varepsilon}$ is constructed via tangent-space PCA (with Riemannian barycentre computed via \cref{eq:thm-bary-remetrized} with $\varphi_{\networkParams}$) using a combination of $400$ samples from the normalizing flow prior $p_{\theta}$ and the $100$ clean reference samples. The latent space dimension is chosen as $d_{\varepsilon} = 40$, giving an effective $\varepsilon \approx 0.04$. We then consider solving inverse problems of the form $\CorVector = \forward \mathbf{x} + \sigma \mathbf{n}$ where $\sigma = 0.05$, $\mathbf{n} \sim \mathcal{N}(\mathbf{0}, \mathbf{I})$, and $\forward$ is the same blur operator used during training, and $\mathbf{x}$ is an unseen digit. We reconstruct images by solving \cref{eq:inverseproblems-loss} with gradient descent with initialization $\mathbf{p}_1 = \mathbf{0}$ and a fixed step size $\alpha = 10^{-2}$. As a baseline, we compare our performance to Total Variation (TV) regularization \cite{rudin1992nonlinear}, where we employ a proximal gradient descent method with iterates \begin{align}
    \mathbf{x}_{t+1} = \mathrm{prox}_{\alpha \lambda \mathrm{TV}}(\mathbf{x}_t - \alpha \forward^\top(\forward \mathbf{x}_t - \CorVector)). \label{eq:tv-alg}
\end{align} For TV, we set $\lambda = 8$ and $\alpha = 0.2/\|\forward\|^2$. For both methods, we choose the reconstruction with the lowest mean square error (MSE). We showcase qualitative results in \cref{fig:inv-probs} on $10$ test images and calculate the average MSE of the reconstruction $\hat{\mathbf{x}}$ with the ground-truth $\mathbf{x}$ via $\mathrm{MSE}:= \frac{1}{d}\|\hat{\mathbf{x}} - \mathbf{x}\|_2^2$.  We see that the RAE decoder prior outperforms the TV algorithm and finds more semantically meaningful reconstructions. 

\section{Conclusion}

In this work, we introduced Riemannian AmbientFlow, a framework for joint manifold learning and generative modeling. To meaningfully reflect on the current method and the directions it opens for future research, it is helpful to discuss the “Riemannian” and “AmbientFlow” components separately, as outlined below. We also highlight potential extensions toward downstream tasks such as geometrically regularized inverse problem solving, where the learned geometric structure may serve as a prior or constraint.





\paragraph{Extensions of the Geometric Setting}
In normalizing flow–based approaches to Riemannian manifold learning, the chosen model for the data distribution inherently defines the assumed underlying geometry and dictates which geometric structures -- such as manifolds -- can be identified and how this extraction is carried out. Consequently, the performance of any such method, including ours, depends on how closely the assumed model distribution aligns with the true data distribution.

Normalizing flows generated by diffeomorphisms with a constant determinant Jacobian may be overly restrictive for realistic data, which often exhibit multimodality, with modes that can be either disjoint or overlapping. Despite promising results on benchmark datasets like MNIST, we believe that advancing toward models based on distributions with non-constant determinant Jacobians will be crucial for capturing geometry and building generative models in the presence of corrupted or complex data -- especially since our recoverability result (\cref{thm:recoverability}) does not impose this restriction.

The key challenge is to ensure that geodesics continue to traverse regions of high data likelihood, a property naturally supported by constant determinant Jacobian architectures. Alongside pursuing more realistic statistical models, progress will depend on developing diffeomorphisms that are simultaneously more expressive and geometrically well-behaved, as well as on designing better mechanisms to enforce low-dimensional structure -- given that, in our experiments, the low-rank regularizer contributed little improvement.


\paragraph{Extensions of the Corruption Setting} Continuing with corruption, in this work, we focused on the setting where each sample was corrupted by a fixed linear operator with different samples of additive noise. Another relevant setting would be in the case where samples may be corrupted by different forward operators $\forward^{(1)},\dots,\forward^{(N)}$. This setting has practical applications such as Very-Long Baseline Interferometry (VLBI) in black hole imaging \cite{akiyama2022first}, where one is continuously generating measurements of an underlying black hole source and the exact location of the telescopes changes over time. This induces different forward operators and noise samples with different statistics. The introduction of different forward operators has also been explored in work on equivariant imaging \cite{chen2023imaging}, where a forward operator can be modified by different group actions. This offers potential benefits to solving the generative modeling problem, as different forward operators may have different nullspaces, potentially providing access to different parts of image space. A more challenging setting of recent interest is in black-box corruption, where the exact forward operator is unknown. This is highly relevant in real-world settings where corruption arising from blur or sensor errors may be difficult to model in an analytical fashion. Finally, it would also be important to address some of the caveats in distributional recovery for \cref{thm:recoverability}, such as analyzing the optimization landscape of problem \cref{eq:rie-ambientflow-problem}.

\paragraph{Pullback Priors for Science and Engineering}
As an application of Riemannian AmbientFlow, we have both theoretically motivated and empirically demonstrated that solving inverse problems through the RAE decoder represents a promising direction for future research. However, the current architectural assumptions -- and, by extension, the geometric assumptions they imply -- are unlikely to capture the full picture. Recent work on optimization over learned data manifolds \cite{diepeveen2025iso} highlights that the interaction between the objective function and the underlying manifold can readily lead to either convex or non-convex optimization landscapes, a nuance that is less immediately apparent within our framework. Future work should therefore aim to characterize potential sources of non-convexity and to establish conditions under which such non-convexity in low-dimensional manifold settings can be effectively mitigated.

\section*{Acknowledgments}
WD is partially supported by the U.S. Department of Energy, Office of Science, Office of Advanced Scientific Computing Research, USA under award DE-SC0025589.

\appendix

\section{Proof of \cref{prop:decoder-smoothness} and Extension}\label{appx:decoder-smoothness}

We first show that for a general diffeomorphism, the decoder $\RAEdecoder_\RAErelerror$ satisfies \cref{def:smoothness-decoder} with explicit constants that depend on the diffeomorphism $\diffeo_\networkParams$ and choice of base point $\bar{\Vector}$.

\begin{proposition}
    Given a smooth diffeomorphism $\diffeo_\networkParams : \R^d \rightarrow \R^d$, consider the affine subspace $S := \diffeo_\networkParams(\bar{\Vector}) + \mathrm{Im}\left(D_{\bar{\Vector}}\diffeo_\networkParams \tangentbasis \right)$ where $\tangentbasis$ has orthonormal columns. Then the RAE decoder $\RAEdecoder_\RAErelerror$ is $(m_1,m_2)$-bi-Lipschitz and its Jacobian is $M$-Lipschitz with the following constants: \begin{align*}
        m_1 & := \inf_{\mathbf{y} \in S} \sigma_{\min}\left(D_{\mathbf{y}}\diffeo_\networkParams^{-1} D_{\bar{\Vector}}\diffeo_\networkParams\tangentbasis\right), \quad
        m_2  := \sup_{\mathbf{y} \in S} \left\|D_{\mathbf{y}}\diffeo_\networkParams^{-1} D_{\bar{\Vector}}\diffeo_\networkParams\tangentbasis\right\|,\ \text{and} \\
        M & := \|D_{\bar{\Vector}}\diffeo_\networkParams\tangentbasis\| \times \sup_{\mathbf{y}_1,\mathbf{y}_2 \in S,\mathbf{y}_1\neq\mathbf{y}_2} \frac{\|D_{\mathbf{y}_1}\diffeo_\networkParams^{-1} - D_{\mathbf{y}_2}\diffeo^{-1}_{\networkParams}\|}{\|\mathbf{y}_1 - \mathbf{y}_2\|_2}.
    \end{align*}
\end{proposition}

\begin{proof}
    The first two bounds immediately follow from the fact that by the Fundamental Theorem of Calculus, for a smooth map $f : \R^r \rightarrow \R^d$, we have for any $\latentVector,\mathbf{q} \in \R^r$ $$f(\latentVector) - f(\mathbf{q})  = \int_0^1J_f(\mathbf{q} + t(\latentVector-\mathbf{q}))(\latentVector-\mathbf{q})\mathrm{d}t.$$ Applying this to the map $\RAEdecoder_\RAErelerror$ and recognizing $J_{\RAEdecoder_\RAErelerror}(\latentVector) = D_{\CorVector(\latentVector)}\diffeo_\networkParams^{-1} D_{\bar{\Vector}}\diffeo_\networkParams\tangentbasis$ where $\CorVector(\latentVector) := \diffeo_\networkParams(\bar{\Vector}) + D_{\bar{\Vector}}\diffeo_\networkParams \mathbf{U}_{\varepsilon} \latentVector$ yields the result. The last result for $M$ follows directly by the sub-multiplicativity of the norm.
\end{proof}

We now focus on proving \cref{prop:decoder-smoothness}, which provides explicit bounds on the bi-Lipschitz and smoothness constants based on the neural network architecture \cref{eq:rae-decoder-definition-tanh-layers}.

\begin{proof}[Proof of \cref{prop:decoder-smoothness}]
    First, note that without loss of generality we assume $\diffeo_{\networkParams}(\bar{\Vector}) = \mathbf{0}$. For notational simplicity, set $\tilde{\mathbf{U}}_{\varepsilon} := D_{\bar{\Vector}}\diffeo_{\networkParams}\mathbf{U}_{\varepsilon}$. Then $\RAEdecoder_\RAErelerror(\latentVector)= \diffeo_{\networkParams}^{-1}(\tilde{\mathbf{U}}_{\varepsilon}\latentVector)$ where \begin{align*}
        \diffeo_{\networkParams}^{-1}(\Vector) = \psi^{(L)} \circ \cdots \circ \psi^{(1)}(\Vector)
    \end{align*} and we define \begin{align*}
        \psi^{(i)}(\Vector) & := (\phi^{(L+1-i)})^{-1}(\Vector) = (\mathbf{V}^{(L+1-i)})^{-1} \circ (f^{(L+1-i})^{-1}(\Vector) =: \tilde{\mathbf{V}}^{(i)} \circ \tilde{f}^{(i)}(\Vector)
    \end{align*} with $\tilde{\mathbf{V}}^{(i)}:=(\mathbf{V}^{(L+1-i)})^{-1}$ and $\tilde{f}^{(i)}(\Vector) := (f^{(L+1-i})^{-1}(\Vector) = [\Vector_1, \Vector_2 - \tilde{g}^{(i)}(\Vector_1)]^\top$ with $\tilde{g}^{(i)}(\mathbf{x}_1) :=g^{(L+1-i)}(\mathbf{x}_1) \in \R^n$. Note that the (upper and lower) Lipschitz constant of $\psi^{(i)}$ is controlled by $\underline{\sigma}_i := \sigma_{\min}(\tilde{\mathbf{V}}^{(i)})$ and $\overline{\sigma}_i = \sigma_{\max}(\tilde{\mathbf{V}}^{(i)})$ along with the largest and smallest singular values of \begin{align*}
        J_{\tilde{f}^{(i)}}(\Vector) = \left[\begin{array}{cc}
            \mathbf{I} & \mathbf{0}  \\
            -J_{\tilde{g}^{(i)}}(\Vector_1) & \mathbf{I}
        \end{array}\right].
    \end{align*} Note that $J_{\tilde{g}^{(i)}}(\Vector_1)$ is diagonal with diagonal entries given by $$\frac{\partial \tilde{g}_{\ell}^{(i)}}{\partial \mathbf{u}_{\ell}}(\mathbf{u}_{\ell}) = \sum_{r=1}^n \alpha^{(L+1-i)}_{r,\ell}r \tanh(\mathbf{u}_{\ell})^{r-1}\mathrm{sech}^2(\mathbf{u}_{\ell}).$$ Since $\tanh(u) \in [0,1)$ and $\mathrm{sech}^2(u) = 1-\tanh^2(u)$, consider the following for integer $r \geq 1$: \begin{align*}
        \sup_{t \in [0,1]} r t^{r-1}(1-t^2).
    \end{align*} When $r = 1$, this is at most $1$. For integer $r \geq 2$, elementary calculus shows that this is maximized when $t_* := \sqrt{(r-1)/(r+1)}$, giving $$\sup_{t \in [0,1]} r t^{r-1}(1-t^2) = \frac{2r}{r+1}\left(\frac{r-1}{r+1}\right)^{\frac{r-1}{2}} \leq 2,\ \forall r \geq 1.$$ This gives $$\left|\frac{\partial \tilde{g}_{\ell}^{(i)}}{\partial \mathbf{u}_{\ell}}(\mathbf{u}_{\ell})\right| \leqslant 2\sum_{r=1}^n |\alpha^{(L+1-i)}_{r,\ell}| =: B^{(i)}_{\ell}\ \text{for every}\ \mathbf{u}_{\ell} \in \R.$$ This guarantees \begin{align*}
        \|J_{\tilde{g}^{(i)}}(\mathbf{u})\| = \max_{\ell \in [d/2]}\left|\frac{\partial \tilde{g}_{\ell}^{(i)}}{\partial \mathbf{u}_{\ell}}(\mathbf{u}_{\ell})\right| \leqslant B_i := \max_{1 \leqslant \ell \leqslant d/2} B^{(i)}_{\ell}.
    \end{align*} Note that due to the block structure of $J_{\tilde{f}^{(i)}}(\Vector)$, we get that $\|J_{\tilde{f}^{(i)}}(\Vector)\| \leqslant 1 + \|J_{\tilde{g}^{(i)}}(\Vector_1)\| \leqslant 1+ B_i.$ Using $J_{\psi^{(i)}}(\Vector) = \tilde{\mathbf{V}}^{(i)}J_{\tilde{f}^{(i)}}(\Vector)$ gives $\|J_{\psi^{(i)}}(\Vector)\| \leqslant (1+B_i)\overline{\sigma}_i$. One can get a lower bound on the smallest singular value of $J_{\tilde{f}^{(i)}}$ by noting that since \begin{align*}
        J_{\tilde{f}^{(i)}}(\Vector)^{-1} = \left[\begin{array}{cc}
            \mathbf{I} & \mathbf{0}  \\
            J_{\tilde{g}^{(i)}}(\Vector_1) & \mathbf{I} 
        \end{array}\right]
        \end{align*} we have that 
        $\|J_{\tilde{f}^{(i)}}(\Vector)^{-1}\| \leqslant 1 + B_i$ which implies $\sigma_{\min}(J_{\psi^{(i)}}) \geqslant \frac{\underline{\sigma}_i}{1+B_i},\ \forall i \in [L].$ Since $\RAEdecoder_\RAErelerror(\latentVector) = \psi^{(L)} \circ \cdots \circ \psi^{(1)}(\tilde{\mathbf{U}}_{\RAErelerror}\latentVector)$, applying the above bounds yields $$m_1 \geq \sigma_{\min}(\tilde{\mathbf{U}}_{\varepsilon})\prod_{i=1}^L\frac{\underline{\sigma}_i}{1+B_i}\ \text{and}\ m_2 \leq \|\tilde{\mathbf{U}}_{\varepsilon}\|\prod_{i=1}^L\overline{\sigma}_i(1+B_i).$$
    
    For the Jacobian-Lipschitz constant, letting $\Vector_{\ell} := \Vector_{\ell}(\latentVector) := \psi^{(\ell)} \circ \dots \circ \psi^{(1)}(\latentVector)$ for $\ell \in [L]$, note that $$J_{D_{\RAErelerror}}(\latentVector) = J_{\psi^{(L)}}(\Vector_{L-1}) J_{\psi^{(L-1)}}(\Vector_{L-2}) \cdots J_{\psi^{(1)}}(\tilde{\mathbf{U}}_{\varepsilon}\latentVector)\tilde{\mathbf{U}}_{\varepsilon}$$ so that $J_{D_{\RAErelerror}}(\latentVector) - J_{D_{\RAErelerror}}(\mathbf{q})$ is equal to \begin{align*}
         \sum_{\ell=1}^LJ_{\psi^{(L)}}(\Vector_{L-1}(\latentVector)) \cdots \left[J_{\psi^{(\ell)}}(\Vector_{\ell-1}(\latentVector)) - J_{\psi^{(\ell)}}(\Vector_{\ell-1}(\mathbf{q}))\right]\cdots J_{\psi^{(1)}}(\tilde{\mathbf{U}}_{\varepsilon}\mathbf{q})\tilde{\mathbf{U}}_{\varepsilon}.
    \end{align*} First note that for each $\ell \in [L]$, we have \begin{align*}
        \|J_{\psi^{(\ell)}}(\Vector) - J_{\psi^{(\ell)}}(\CorVector)\|
        & \leqslant \left\|\tilde{\mathbf{V}}^{(\ell)}(J_{\tilde{f}^{(\ell)}}(\Vector)-J_{\tilde{f}^{(i)}}(\CorVector)) \right\|\\
        & \leqslant\|\tilde{\mathbf{V}}^{(\ell)}\|\|J_{\tilde{g}^{(\ell)}}(\CorVector_1) -J_{\tilde{g}^{(\ell)}}(\Vector_1)\| \\
        & \leqslant \overline{\sigma}_{\ell}\mathrm{Lip}(J_{\tilde{g}^{(\ell)}})\|\CorVector-\Vector\|_2
    \end{align*} where $\mathrm{Lip}(J_{\tilde{g}^{(\ell)}})$ is the Lipschitz constant of the Jacobian $J_{\tilde{g}^{(\ell)}}$. To estimate this, note that \begin{align*}
        \frac{\partial^2 \tilde{g}^{(\ell)}_i}{\partial \mathbf{u}_i}(\mathbf{u}_i) = \sum_{r=1}^n &\alpha^{(L+1-\ell)}_{r,i}r\big[(r-1)\tanh(\mathbf{u}_i)^{r-2}\mathrm{sech}^4(\mathbf{u}_i) - 2\tanh(\mathbf{u}_i)^r\mathrm{sech}^2(\mathbf{u}_i)\big].
    \end{align*} When $r = 1$, the term is bounded by $|\alpha^{(L+1-\ell)}_{r,i}|$. For integer $r \geq 2$, we can upper bound this by \begin{align*}
        \sup_{t \in [0,1)} rt^{\frac{r-2}{2}}(1-t)|(r-1)-(r+1)t|.
    \end{align*} Elementary calculus shows that this quantity is always bounded by $2$ for any integer $r \geq 2$, which gives \begin{align*}
       \mathrm{Lip}(J_g^{(\ell)}) \leq 2\max_{i} \sum_{r=1}^n|\alpha^{(L+1-\ell)}_{r,i}| =: \tilde{C}_{\ell}.
    \end{align*} Using these bounds and letting $\tilde{M}_i := (1+B_i)\overline{\sigma}_i$, we get that \begin{align*}
        \|J_{D_{\varepsilon}}(\latentVector)-J_{D_{\varepsilon}}(\mathbf{q})\| \leqslant \|\tilde{\mathbf{U}}_{\varepsilon}\|\sum_{\ell=1}^L \tilde{M}_{L}\dots \tilde{M}_{\ell+1}\overline{\sigma}_i\mathrm{Lip}(J_g^{(\ell)})\|\Vector_{\ell-1}(\latentVector)-\Vector_{\ell-1}(\mathbf{q})\|_2\tilde{M}_{\ell-1}\dots\tilde{M}_{1}.
    \end{align*} Finally, we note that $\|\Vector_{\ell-1}(\latentVector) - \Vector_{\ell-1}(\mathbf{q})\| \leqslant \prod_{i=1}^{\ell-1}\tilde{M}_i\|\latentVector-\mathbf{q}\|_2$ which gives the final bound: \begin{align*}
        \|J_{D_{\RAErelerror}}(\latentVector)-J_{D_{\RAErelerror}}(\mathbf{q})\| \leqslant \|\tilde{\mathbf{U}}_{\varepsilon}\|\sum_{\ell=1}^L\prod_{i=\ell+1}^L\tilde{M}_i\overline{\sigma}_i\tilde{C}_{\ell}\left(\prod_{k=1}^{\ell-1}\tilde{M}_k\right)^2\|\latentVector-\mathbf{q}\|_2.
    \end{align*}
\end{proof}

\bibliographystyle{plain}

\bibliography{references}

\end{document}